%% file: neurips_2026.tex
\documentclass{article}

\PassOptionsToPackage{square, numbers}{natbib}

 \usepackage[main, final]{neurips_2026}
\usepackage[utf8]{inputenc} % allow utf-8 input
\usepackage[T1]{fontenc}    % use 8-bit T1 fonts
\usepackage{hyperref}       % hyperlinks
\usepackage{url}            % simple URL typesetting
\usepackage{booktabs}       % professional-quality tables
\usepackage{amsfonts}       % blackboard math symbols
\usepackage{nicefrac}       % compact symbols for 1/2, etc.
\usepackage{microtype}      % microtypography
\usepackage{xcolor}         % colors
\usepackage{graphicx}  

\usepackage{subcaption}
\usepackage{amsmath}
\usepackage{enumitem}
\usepackage{wrapfig}

\newcommand{\eg}{\emph{e.g., }}

\title{Retrieval-Augmented Diffusion Modeling for Stochastic Discount Factor Portfolios}
\vspace{-5px}

\author{
Kelvin J.L. Koa\textsuperscript{*\textdagger}\;
Xinyang Li\textsuperscript{*}\;
Ke-Wei Huang\;
\\
National University of Singapore \quad
Asian Institute of Digital Finance \quad
\\
\texttt{kelvin.koa@u.nus.edu, xinyangli5579@gmail.com, dishkw@nus.edu.sg}
}

\begin{document}

\maketitle

\begingroup
\renewcommand\thefootnote{}
\footnotetext{* Equal contribution. \ \textdagger\ Corresponding author.}
\endgroup

\vspace{-6px}
\begin{abstract}
\vspace{-6px}
In this work, we study portfolio optimization under the stochastic discount factor (SDF) framework by learning market state representations that capture the underlying risk structures of financial data. This is challenging due to several factors: financial markets exhibit non-stationary dynamics with shifting regimes, multimodal inputs such as price and news data often contain stochastic noise, and existing diffusion-based approaches, while effective for modeling stochastic dynamics, rely on assumptions such as isotropic Gaussian noise that fail to capture the state-dependent nature of financial uncertainty. To address these challenges, we introduce \textsc{radar}, a retrieval-augmented diffusion framework that learns market representations by conditioning on similar historical regimes. \textsc{radar} leverages retrieval to construct context-dependent noise distributions, applies conditional diffusion to denoise multimodal representations, and initializes the diffusion process using empirical statistics to reflect state-dependent uncertainty. Experiments show that \textsc{radar} achieves state-of-the-art performance on key risk-adjusted metrics while producing economically meaningful signals on asset returns and correlations. Our code is available at: \url{https://github.com/xinyangli5579-star/RADAR}.
% In this work, we study portfolio optimization under the stochastic discount factor (SDF) framework. To do this, we learn market state representations that can capture the underlying risk structures of financial data. This is challenging due to several factors: financial markets exhibit non-stationary dynamics with shifting regimes, multimodal inputs such as price and news data often contain stochastic noise, and standard diffusion model assumptions such as isotropic Gaussian noise fail to capture the state-dependent nature of financial uncertainty. To address these, we introduce \textsc{radar}, a retrieval-augmented diffusion framework that learns market representations by conditioning on similar historical regimes. \textsc{radar} leverages retrieval to construct context-dependent noise distributions, applies conditional diffusion to denoise multimodal representations, and initializes the diffusion process using empirical statistics to reflect state-dependent uncertainty. Experiments show that \textsc{radar} achieves state-of-the-art performance on key risk-adjusted metrics while producing economically meaningful signals on asset returns and correlations.
\vspace{-11px}
\end{abstract}

\input{1_introduction_updated}
\input{2_related}
\input{3_model}
\input{4_experiments}
\input{5_results}
\input{6_conclusion}

\newpage
\bibliographystyle{plainnat}
\bibliography{reference}

\appendix
\newpage
\input{a_proof_final}
\input{b_appendix}

% \newpage
% \input{checklist.tex}

\end{document}

%% file: 1_introduction_updated.tex
\section{Introduction} 
\vspace{-3px}
Portfolio optimization is a central problem in quantitative finance, where the objective is to allocate capital across assets to achieve an optimal trade-off between risk and return \cite{markowitz1952utility}. Classical approaches such as mean–variance optimization \cite{markowitz1952portfolio} formalize this trade-off by maximizing expected returns subject to a penalty on its variance, thereby optimizing for stable portfolio returns under uncertain conditions. In modern asset pricing theory, this objective can be equivalently expressed through the stochastic discount factor (SDF) \cite{kelly2024large, kelly2025artificial}, which provides a unified framework for valuing uncertain future financial payoffs. The SDF characterizes asset values as expectations under a market state-dependent weighting of payoffs, thereby encoding both risk and return within a single object.
\begin{figure*}[h]
\vspace{-1px}
    \centering
    \begin{subfigure}[t]{0.32\textwidth}
        \centering
        \includegraphics[width=0.77\linewidth]{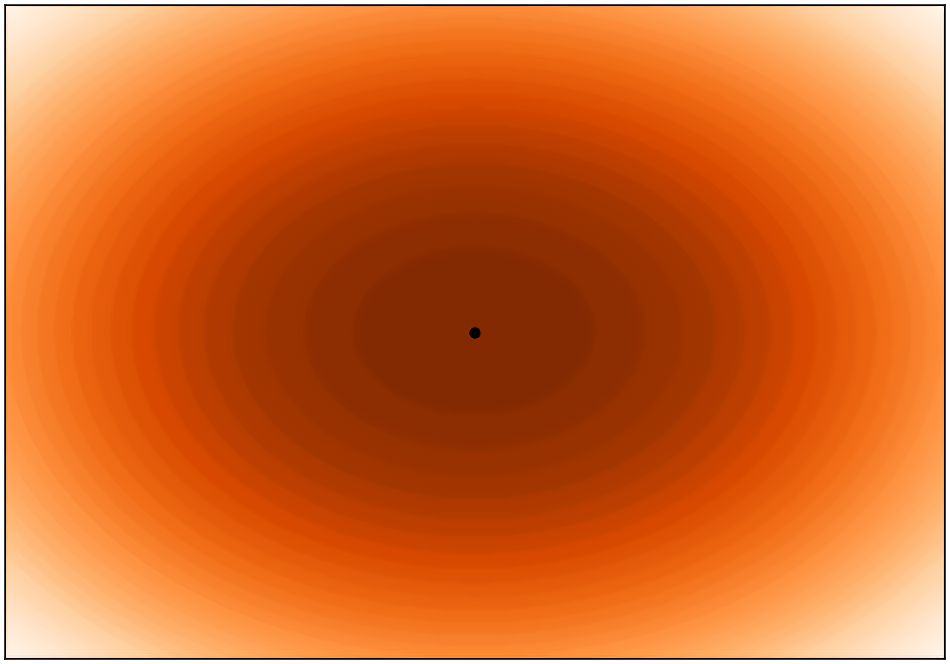}
        \caption{The returns forecasting task optimizes for a single point optimum.}
        \label{fig:forecasting_task}
    \end{subfigure}
    \hfill
    \begin{subfigure}[t]{0.32\textwidth}
        \centering
        \includegraphics[width=0.77\linewidth]{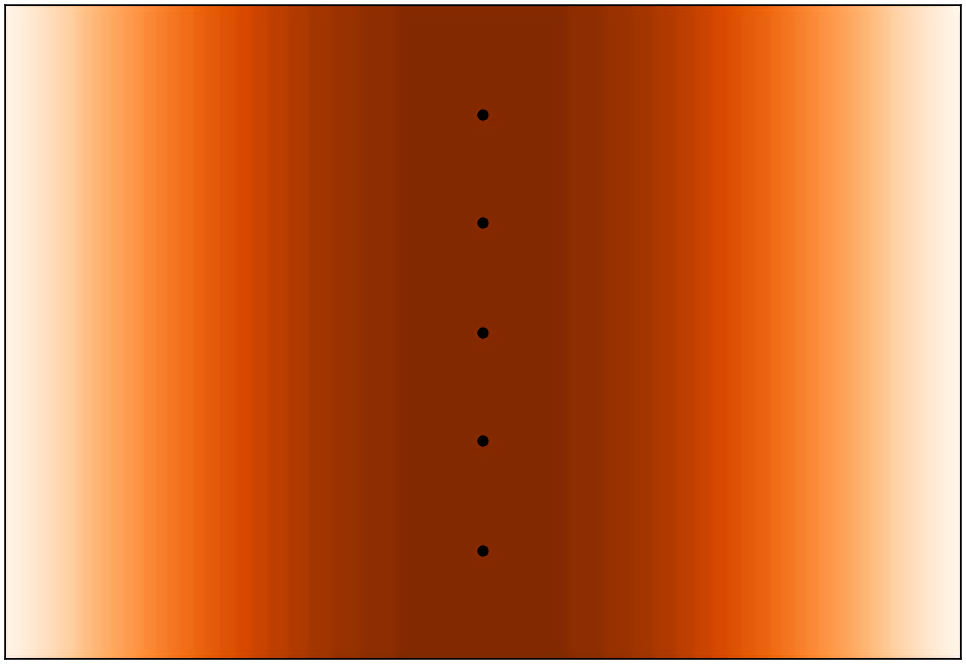}
        \caption{The SDF estimation task admits a manifold of equivalent solutions.}
        \label{fig:sdf_task_clean}
    \end{subfigure}
    \hfill
    \begin{subfigure}[t]{0.32\textwidth}
        \centering
        \includegraphics[width=0.77\linewidth]{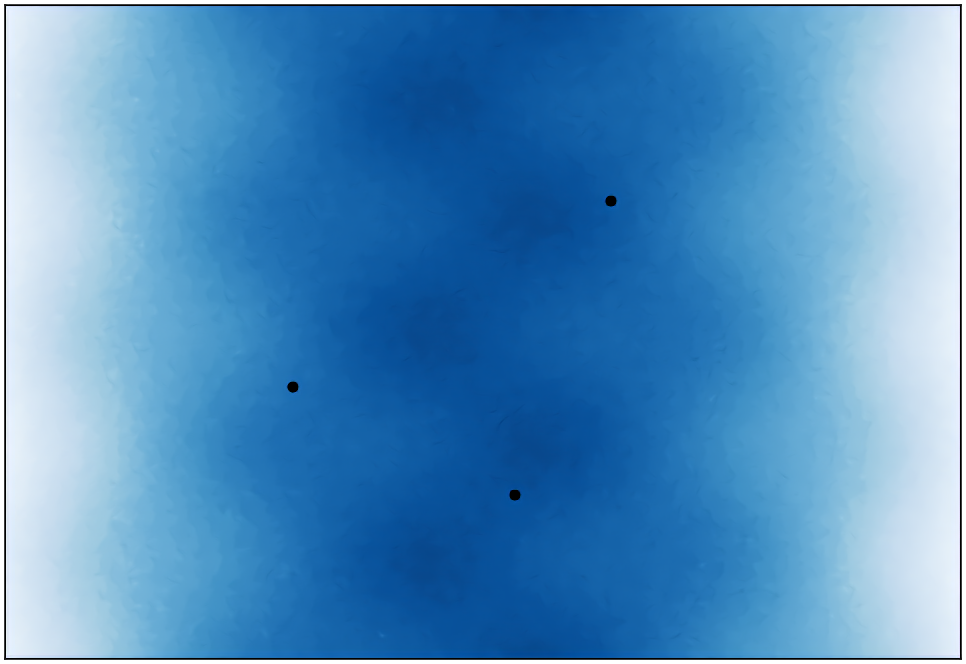}
        \caption{Noisy representations distort manifold, causing spurious minima.}
        \label{fig:sdf_task_noisy}
    \end{subfigure}
    \vspace{-2px}
    \caption{
    Comparison of forecasting and SDF objectives, and the impact of noisy representations.
    }
    \label{fig:forecasting_vs_sdf_vs_noisy}
    \vspace{-7px}
\end{figure*}

% In parallel, advances in deep learning have also led to a range of data-driven approaches for modeling financial markets. For example, prior works in financial forecasting have explored using news data  to capture market trends and volatility \cite{hu2018listening, xu2018stock}, or using diffusion noise to model the stochasticity of asset prices \cite{koa2023diffusion, gao2024diffsformer}. However, these methods are primarily optimized for the returns forecasting task. 
Unlike returns forecasting \cite{ding2015deep}, which focuses on predicting conditional expectations, SDF estimation is inherently underdetermined and requires learning representations that are sufficient to satisfy pricing constraints across all assets (see Figure \ref{fig:forecasting_vs_sdf_vs_noisy}\subref{fig:forecasting_task}-\subref{fig:sdf_task_clean}). This requires capturing the joint structure of market states and their associated risk, rather than extracting predictive signals for individual assets.

To obtain market representations, two fundamental limitations exist in current deep-learning methods. Firstly, financial markets are inherently non-stationary, often exhibiting regime shifts \cite{hamilton1989new} that alter the structural relationships between features and returns. As a result, models can be biased by the relative coverage of different regimes in the dataset. Some methods attempt to address non-stationarity through temporal weighting \cite{liu2022non} or trend decomposition \cite{wu2021autoformer}, but they typically rely on heuristics and fail to capture contextual similarities across market states. Secondly, while textual data provides useful signals \cite{hu2018listening, xu2018stock}, it is often noisy and contains irrelevant information with limited economic relevance \cite{tetlock2007giving}. Training on such embeddings makes it harder for the model to capture economically meaningful signals, leading to suboptimal pricing and asset allocation decisions (see Figure \ref{fig:forecasting_vs_sdf_vs_noisy}\subref{fig:sdf_task_noisy}).

Among existing modeling paradigms, diffusion-based methods have recently emerged as a promising approach for learning robust representations and capturing uncertainty in stochastic environments \cite{koa2023diffusion, gao2024diffsformer}. However, they typically operate only on the time-series modality and rely on isotropic Gaussian noise during training, which does not capture the state-dependent and heteroskedastic nature of financial uncertainty. The closest alternative is  Retrieval-Augmented Time-series Diffusion model (RATD) \cite{liu2024retrieval}, which retrieves similar historical states as reference to a diffusion model. However, these retrieved states are used to condition the denoising process rather than guide the stochastic noise generation in the diffusion process. Furthermore, they also do not deal with multimodal inputs.  

\begin{figure*}[t]
    \vspace{-15px}
    \centering
    \includegraphics[width=0.88\textwidth]{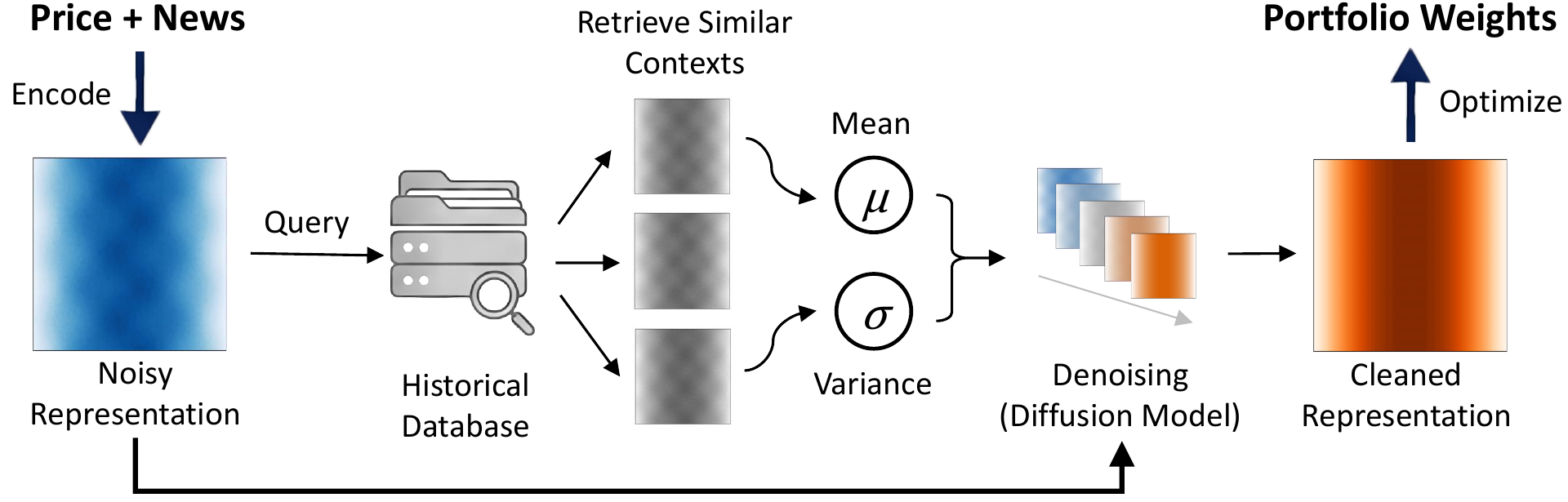}
    \caption{
    Overview of \textsc{radar}, which learns a clean market state representation for SDF estimation.
    }
    \label{fig:radar_overview}
    \vspace{-10px}
\end{figure*}

To address these, we propose \textbf{R}etrieval-\textbf{A}ugmented \textbf{D}iffusion-based \textbf{A}ssets \textbf{R}epresentation (\textsc{radar}), a framework for learning market representations for SDF estimation (see Figure \ref{fig:radar_overview}). Firstly, to account for non-stationarity, we introduce a retrieval-augmented diffusion process that conditions on contextually similar historical regimes. This enables regime-aware representations that are grounded in relevant historical conditions rather than aggregated across unrelated periods. Secondly, to mitigate noise in unstructured data, we treat the latent representations of both price and news as corrupted observations and refine them via a conditional diffusion process, which iteratively denoises the representations. This allows the model to generalize across similar contexts while suppressing the idiosyncratic noise from individual observations. Lastly, to capture state-dependent uncertainty, we initialize the diffusion process using the mean and variance of the retrieved contexts. This grounds the denoising trajectory in empirical conditional distributions, enabling the learned representations to better reflect the risk–return characteristics of the current market state for downstream SDF estimation.

% To demonstrate the effectiveness of \textsc{radar}, we perform extensive experiments across forecasting models and portfolio optimization methods to assess the effectiveness of \textsc{radar}. Our results show that \textsc{radar} consistently achieves state-of-the-art performance across key risk-adjusted metrics, including Sharpe, Sortino, and Calmar ratios, indicating a superior risk-return trade-off. Through ablation studies, we demonstrate the importance of representation learning and context-aware diffusion in driving performance gains. Finally, additional analyses on the learned representations show that the model produces economically meaningful signals that translate into improved portfolio outcomes.
Experimental results show that \textsc{radar} consistently achieves state-of-the-art performance across key risk-adjusted metrics, including Sharpe, Sortino, and Calmar ratios. 
% Through ablation studies, we demonstrate the importance of representation learning and context-aware diffusion in driving performance gains. Finally, 
Additional analyses on the learned representations also show that the model produce economically meaningful signals on asset returns and correlations that are beneficial for portfolio outcomes.
In summary, our contributions are:
\begin{itemize}[leftmargin=*]
\item We introduce a state-dependent diffusion formulation where the noise distribution is conditioned on retrieved historical contexts, addressing a key mismatch between standard diffusion assumptions and financial data.
\item We show that retrieval can be used to model conditional uncertainty, enabling regime-aware representation learning in non-stationary environments.
\item We propose the \textsc{radar} framework for learning clean market representations for SDF-based portfolio optimization using multimodal inputs and context-aware denoising.
\item We demonstrate strong empirical performance and provide analyses showing that the learned representations are economically meaningful and effective for portfolio construction.
\end{itemize}

%% file: 2_related.tex
\section{Related Works}

\textbf{Financial Forecasting.}
The financial forecasting task is primarily concerned with extracting predictive signals from input data to estimate stock prices. While early approaches in financial forecasting explored the applicability of deep learning techniques on forecasting stock movements \cite{ding2015deep, qin2017dual}, recent works have seen practitioners incorporating stylized financial characteristics into their model designs. For example, works have explored the use of textual information \cite{hu2018listening, xu2018stock} to capture market trends and volatility, following the concept of informationally efficient financial markets \cite{fama1970efficient, koa2024learning}. Other works have also explored using diffusion noise to model the stochastic noise in financial prices \cite{koa2023diffusion, gao2024diffsformer}, following the idea of randomness in financial markets \cite{malkiel1973random, feng2018enhancing}. These two perspectives are typically studied in isolation in existing works. In contrast, our approach integrates both information-driven signals and stochastic noise modeling, and applies this unified framework to portfolio applications.

\textbf{Deep Learning for Portfolio Optimization.}
The portfolio optimization task aims to allocate capital across multiple assets to achieve optimal balance between risk and returns \cite{markowitz1952utility}. Beyond classical mean-variance approaches \cite{markowitz1952portfolio}, recent works have also seen practitioners incorporate deep-learning methods into portfolio optimization applications. For example, Hierarchical Risk Parity (HRP) \cite{lopez2016building} leverages hierarchical clustering to construct diversified portfolios based on graph theory; reinforcement learning (RL) is used to learn portfolio weights that dynamically adjust over time \cite{jiang2017deep, wang2019alphastock}. The most recent theory-based approaches adopt the stochastic discount factor (SDF) framework \cite{kelly2024large, kelly2025artificial}, which provides a unified formulation for evaluating assets through risk-adjusted valuation. Rather than directly forecasting returns, this framework learns a pricing model that leverages shared structure across assets, implicitly capturing risk premia and state-dependent risk-return trade-offs. This aligns with a broader trend of learning structured representations in financial markets \cite{sarmah2024learning, mehta2025clustering, gabaix2025asset} which aim to encode cross-asset relationships for downstream portfolio decisions. However, learning such asset representations remains challenging in practice due to non-stationarity, noisy inputs, and the context-dependent nature of financial risk, which our work aims to tackle.

\textbf{Non-Isotropic Diffusion Processes.}
Standard diffusion models typically adopt a fixed isotropic Gaussian noise process \cite{ho2020denoising}. However, recent works have also explored the use of non-isotropic diffusion processes. For example, some approaches learn input-dependent multivariate noise schedules, allowing different data dimensions to receive noise at different rates \cite{sahoo2024diffusion}. Other works explored using flexible forward processes to make the reverse generation path easier to learn \cite{bartosh2024neural}, or performing time-dependent non-linear transformations of the data during diffusion \cite{bartosh2023neural}. In time-series forecasting, works have introduced learned non-linear transformations and conditioning variables into the forward process \cite{rishi2025conditional}, or adapted the endpoint distribution and noise schedule using forecasted conditional statistics \cite{ye2025nonstationary, li2024transformer}. However, most of these works primarily learn how to modify the diffusion process, whereas our approach constructs the noise distribution directly from retrieved historical samples.

%% file: 3_model.tex
\section{The \textsc{radar} Framework}
\begin{figure*}[t]
    \centering
    \includegraphics[width=\textwidth]{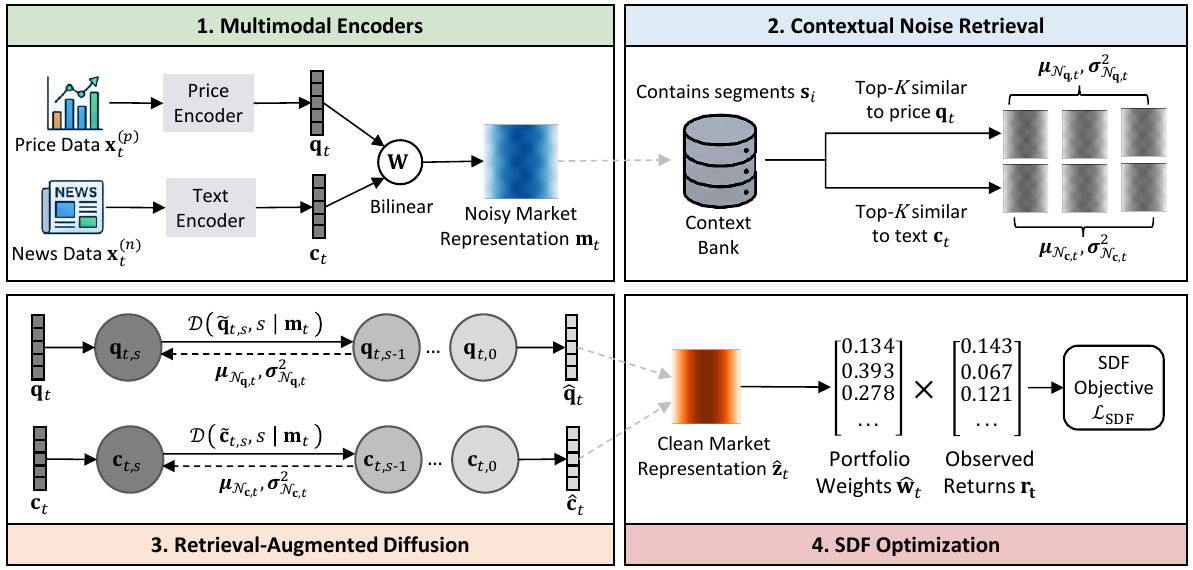}
    \caption{
    The \textsc{radar} framework. The price and text embeddings, noisy market representation and retrieved contextual noise are used in the diffusion process to obtain the clean market representation.
    }
    \label{fig:radar_diagram}
\end{figure*}

In this section, we first study the portfolio optimization problem within the stochastic discount factor (SDF) framework from a deep-learning perspective. We then explain the proposed \textbf{R}etrieval-\textbf{A}ugmented \textbf{D}iffusion-based \textbf{A}ssets \textbf{R}epresentation (\textsc{radar}) framework, shown in Figure \ref{fig:radar_diagram}. 

\subsection{Problem Formulation}
At each time step $t$, we observe two input modalities which are describing the market state: time-series asset prices $\mathbf{x}^{(p)}_t$ and textual news data $\mathbf{x}^{(n)}_t$. 
Our goal is to learn the joint representative embeddings: 
\begin{equation}
\hat{\mathbf{z}}_t = f(\mathbf{x}^{(p)}_t, \mathbf{x}^{(n)}_t),
\end{equation}
which should capture useful information about the current market state. This representation is used to learn the asset portfolio weights $\hat{\mathbf{w}}_t \in \mathbb{R}^{N}$, where $N$ is the number of assets. Formally, we have:
\begin{equation}
\hat{\mathbf{w}}_t = g(\hat{\mathbf{z}}_t).
\end{equation}
% We adopt the stochastic discount factor (SDF) framework \cite{kelly2025artificial, hansen1987role} to define a principled training objective. Given the next-period asset return $\mathbf{r}_{t+1} \in \mathbb{R}^{N}$, the predicted weights induce a portfolio return $\hat{\mathbf{w}}_t^\top \mathbf{r}_{t+1}$, where a value of $1$ corresponds to no change in wealth. The SDF is then defined as $1 - \hat{\mathbf{w}}_t^\top \mathbf{r}_{t+1}$, which measures the deviation of the realized portfolio outcome from this baseline. 
We adopt the stochastic discount factor (SDF) framework \cite{kelly2025artificial, hansen1987role} to define a principled training objective. Given the next-period asset return $\mathbf{r}_{t+1} \in \mathbb{R}^N$, the predicted weights induce a portfolio return $\hat{\mathbf{w}}_t^\top \mathbf{r}_{t+1}$, where a value of 0 corresponds to no change in wealth. The SDF is then defined as $1 - \hat{\mathbf{w}}_t^\top \mathbf{r}_{t+1}$, and minimizing its second moment makes it price every asset with zero error \cite{kelly2025artificial}.

% Intuitively, minimizing this deviation encourages the model to ensure that portfolio outcomes are consistent with the information used to construct them. Both positive and negative deviations from the baseline indicate a mismatch between the learned representation and future returns. In particular, systematic deviations suggest that the representation $\hat{\mathbf{z}}_t$ fails to capture structure in the returns that could be explained by the input data. By penalizing such deviations, the model is driven to learn representations that account for the predictable components of returns, rather than spurious patterns.
Intuitively, minimizing this pricing error encourages the model to ensure that portfolio outcomes are consistent with the information used to construct them. Both positive and negative pricing errors on an asset indicate a mismatch between the learned representation and future returns. In particular, systematic mispricing suggests that the representation $\hat{\mathbf{z}}_t$ fails to capture structure in the returns that could be explained by the input data. By penalizing such errors, the model is driven to learn representations that account for the predictable components of returns, rather than spurious patterns.

Following this, we then train the model by minimizing the corresponding SDF-based objective:
\begin{equation}
\mathcal{L}_{\mathrm{SDF}}
=
\mathbb{E}_t
\left[
\left(1 - \hat{\mathbf{w}}_t^\top \mathbf{r}_{t+1}\right)^2
\right]
+
\lambda \|\hat{\mathbf{w}}_t\|_2^2,
\label{eq:sdf_loss}
\end{equation}
where $\lambda \|\hat{\mathbf{w}}_t\|_2^2$ is a $\ell_2$ regularization term to prevent any unstable extreme \cite{ledoit2003honey} portfolio weights.

\subsection{Multimodal Encoders}
To obtain the initial embeddings from the heterogeneous multi-modal inputs, we first employ modality-specific encoders \cite{sawhney2020deep} for the time-series asset prices and textual news data, and blend them.

For the price time-series data, we model the temporal dependencies using a gated recurrent unit (GRU) price encoder. Given the historical price sequence $\mathbf{x}^{(p)}_t$, we extract the hidden states, which are then aggregated via a temporal attention mechanism $\mathrm{Attn}_p$ to produce the price representation:
\begin{equation}
\mathbf{q}_t = \mathrm{Attn}_p\big(\mathrm{GRU}_p(\mathbf{x}^{(p)}_t)\big),
\label{eq:price_encoder}
\end{equation}
For the text modality, we leverage pre-trained financial language models \cite{araci2019finbert} to extract embeddings from the textual news data $\mathbf{x}_t^{(n)}$. We then model the sequential dependencies of these embeddings using a GRU news encoder and a sequential attention mechanism to derive the text representation:
\begin{equation}
\mathbf{c}_t = \mathrm{Attn}_n\left(\mathrm{GRU}_n \big(\mathrm{Embed}(\mathbf{x}_t^{(n)})\big)\right),
\label{eq:text_encoder}
\end{equation}
Given the modality-specific representations $\mathbf{q}_t$ and $\mathbf{c}_t$, we blend them by applying a bilinear transformation which captures cross-modality dependencies. Specifically, we compute a joint representation:
\begin{equation}
\mathbf{m}_t = \mathrm{ReLU}\left(\mathbf{q}_t^\top \mathbf{W} \mathbf{c}_t + \mathbf{b}\right),
\label{eq:multimodal_fusion}
\end{equation}
where $\mathbf{W}$ and $\mathbf{b}$ are the learnable parameters, and $\mathrm{ReLU}$ is the non-linear activation function.

\subsection{Contextual Noise Retrieval}
The learnt representation is unstable, as asset price data contains stochastic noise \cite{koa2023diffusion} while news data often contains chaotic irrelevant information \cite{hu2018listening}, which requires denoising. On the other hand, standard diffusion denoising techniques typically rely on isotropic Gaussian noise, which is independent of the input and may fail to preserve meaningful structure in the representation.

To address this issue, we introduce a retrieval-based mechanism to construct context-dependent noise distributions from historically similar market states. The data is first organized around news events. Let $\{\tau_i\}_{i=1}^{M}$ denote the time indices at which news events arrive, ordered chronologically. For each news event at time $\tau_i$, we define an event segment consisting of the news observed at that time together with their subsequent asset price sequences up to, but excluding, the next news event:
\begin{equation}
\mathbf{s}_i =
\left(
\mathbf{x}^{(n)}_{\tau_i},\;
\mathbf{x}^{(p)}_{\tau_i:\tau_{i+1}-1}
\right).
\label{eq:segment}
\end{equation}
Each segment $\mathbf{s}_i$ captures the impact of a news event on the market before new information arrives. We then encode each event segment using the same encoders and bilinear transformation used in Equations \ref{eq:price_encoder}, \ref{eq:text_encoder} and \ref{eq:multimodal_fusion}, to obtain the segment representations $\tilde{\mathbf{s}}_i$, which are stored in a \textit{context bank}.

At each time step $t$, we retrieve the most relevant historical segments in the current context bank across both modalities, calculated using the cosine similarity metric. For each modality, we have:
\begin{equation}
\mathcal{N}_{\mathbf{q},t} =
\operatorname{TopK}_{i:\tau_i < t}
\cos\big(\mathbf{q}_t, \tilde{\mathbf{s}}_i\big),
\label{eq:retrieval_q}
\end{equation}
This captures similar historical price trajectories, and the news information that just preceded them.
% For the text modality, we have:
\begin{equation}
\mathcal{N}_{\mathbf{c},t} =
\operatorname{TopK}_{i:\tau_i < t}
\cos\big(\mathbf{c}_t, \tilde{\mathbf{s}}_i\big),
\label{eq:retrieval_c}
\end{equation}
This captures historical news with similar semantic content, and their subsequent price trajectories.

The retrieved sets represent the possible states that the market could be in, for each modality. We model their means and variances as noise distribution parameters. For the price modality, we have:
\begin{equation}
\boldsymbol{\mu}_{\mathcal{N}_{\mathbf{q},t}} = \frac{1}{K} \sum_{k=1}^{K} \tilde{\mathbf{s}}_{k}, \quad
\boldsymbol{\sigma}^{2}_{\mathcal{N}_{\mathbf{q},t}} = \frac{1}{K} \sum_{k=1}^{K} \left(\tilde{\mathbf{s}}_{k} - \boldsymbol{\mu}_{\mathcal{N}_{\mathbf{q},t}}\right)^2,
\qquad
\tilde{\mathbf{s}}_{k} \in \mathcal{N}_{\mathbf{q},t}.
\label{eq:retrieval_stats_q}
\end{equation}
The noise distribution of the text modality $\left(\boldsymbol{\mu}_{\mathcal{N}_{\mathbf{c},t}}, \boldsymbol{\sigma}^{2}_{\mathcal{N}_{\mathbf{c},t}}\right)$ is then calculated in a similar manner.

\subsection{Retrieval-Augmented Diffusion}
In this step, we refine the learnt modality embeddings $\mathbf{q}_t$ and $\mathbf{c}_t$ through a diffusion process, guided by the retrieved contextual noise distribution and conditioned on the joint market representation.

\paragraph{Forward Process.}
We adopt a dual-path formulation that perturbs the price and text representations independently. At each diffusion step $s$, we inject the retrieved contextual noise into the embeddings:
\begin{align}
\tilde{\mathbf{q}}_{t,s}
&= \sqrt{\bar{\alpha}_s}\mathbf{q}_t
+ \sqrt{1-\bar{\alpha}_s}
\left(
\boldsymbol{\mu}_{\mathcal{N}_{\mathbf{q},t}}
+ \boldsymbol{\sigma}_{\mathcal{N}_{\mathbf{q},t}}
\odot \boldsymbol{\epsilon}_{\mathbf{q}}
\right),
\label{eq:forward_q}
\\
\tilde{\mathbf{c}}_{t,s}
&= \sqrt{\bar{\alpha}_s}\mathbf{c}_t
+ \sqrt{1-\bar{\alpha}_s}
\left(
\boldsymbol{\mu}_{\mathcal{N}_{\mathbf{c},t}}
+ \boldsymbol{\sigma}_{\mathcal{N}_{\mathbf{c},t}}
\odot \boldsymbol{\epsilon}_{\mathbf{c}}
\right),
\label{eq:forward_c}
\end{align}
where $\boldsymbol{\epsilon}_\mathbf{q}, \boldsymbol{\epsilon}_\mathbf{c} \sim \mathcal{N}(\mathbf{0}, \mathbf{I})$, and $\bar{\alpha}_s$ controls the noise schedule.
Unlike standard diffusion models that employ isotropic Gaussian noise, the perturbations are governed by context-specific statistics, allowing them to model potential unrealized paths using historically similar market states.

\paragraph{Reverse Process.}
To recover clean representations, we employ the denoising functions $\mathcal{D}_p$ and $\mathcal{D}_n$:
\begin{equation}
\hat{\mathbf{q}}_t = \mathcal{D}_p\big(\tilde{\mathbf{q}}_{t,s}, s \mid \mathbf{m}_t\big),
\qquad
\hat{\mathbf{c}}_t = \mathcal{D}_n\big(\tilde{\mathbf{c}}_{t,s}, s \mid \mathbf{m}_t\big).
\label{eq:denoising}
\end{equation}
Each denoiser conditions on the shared market representation $\mathbf{m}_t$, enabling cross-modal information exchange during reconstruction, while denoising towards a generalized state for each modality.

The denoiser contains a self-attention layer, applied to the two noisy embeddings $\mathbf{q}_t$ and $\mathbf{c}_t$, together with the market-state conditioning $\mathbf{m}_t$, and a sinusoidal embedding of the diffusion step $s$. The layer applies LayerNorm, followed by the query, key, value projections and scaled dot-product attention. 

\paragraph{Denoising Objective.}
We train the model using a joint reconstruction objective over both modalities:
\begin{equation}
\mathcal{L}_{\mathrm{Diff}} = \mathbb{E}_{t,s} \left[
\left\| \mathbf{q}_t - \hat{\mathbf{q}}_t \right\|_2^2 +
\left\| \mathbf{c}_t - \hat{\mathbf{c}}_t \right\|_2^2
\right].
\label{eq:diff_loss}
\end{equation}
This objective encourages the model to reconstruct modality-specific representations from their noisy counterparts, serving as refined features for subsequent market representation construction.

A derivation of the diffusion process formulation used in \textsc{radar} can be found in Appendix \ref{app:diffusion_formulation}.

\paragraph{Final Representation.}
Using the denoised modality-specific representations $\hat{\mathbf{q}}_t$ and $\hat{\mathbf{c}}_t$, we construct the final, refined market representation by aggregating across the refined cross-modal signals:
\begin{equation}
\hat{\mathbf{z}}_t
= (1-\gamma)\left(\hat{\mathbf{q}}_t + \hat{\mathbf{c}}_t\right)
+ \gamma\mathbf{q}_t,
\label{eq:final_representation}
\end{equation}
where $\gamma$ is a hyperparameter that balances the contribution between the denoised cross-modal information and the original price representation. The price representation is used as an additional signal for portfolio optimization, with cross-modal information serving as the cleaned state representation.

\subsection{Overall Training Objective}
Finally, the model is trained by jointly optimizing the diffusion and SDF-based portfolio objective:
\begin{equation}
\mathcal{L} = \mathcal{L}_{\mathrm{SDF}} + \rho\cdot\mathcal{L}_{\mathrm{Diff}},
\end{equation}
where $\rho$ balances the market representation learning and downstream portfolio optimization task.

%% file: 4_experiments.tex
\section{Experiments} \label{sec:experiments}
\paragraph{Dataset.} \label{para:dataset} We evaluate \textsc{radar} on the S\&P 500 constituent stocks. We collect daily adjusted close prices from Yahoo Finance and financial news articles from FNSPID \cite{dong2024fnspid}, and the final dataset includes 377 U.S. equities that appear in both the S\&P 500 index and the news corpus. The duration spans from June 2011 to June 2020, covering approximately 2,270 trading days. To avoid look-ahead bias, we adopt a rolling-window protocol: each window uses 4 years of data for training (with a 90:10 train/validation split) and the subsequent 1 year for out-of-sample testing, rolling forward by 1 year.

\paragraph{Baselines.} Following our motivation, we compare \textsc{radar} against two families of baselines.

% \textit{Time-series forecasting models.} We compare against StockNet \cite{xu2018stock}, which incorporates both price and textual signals; HAN \cite{hu2018listening}, which models hierarchical text representations; NGAT \cite{niu2025ngat}, which captures cross-asset relationships via graph attention; and iTransformer \cite{liu2023itransformer}, a recent transformer-based time-series model. These models predict future stock returns, from which we construct portfolios by ranking the predicted returns and equally weighting the top-50 stocks at each rebalancing date.
\textit{Time-series forecasting models.} We compare against StockNet \cite{xu2018stock}, which incorporates both price and textual signals; HAN \cite{hu2018listening}, which models hierarchical text representations; NGAT \cite{niu2025ngat}, which captures cross-asset relationships via graph attention; iTransformer \cite{liu2023itransformer}, a recent transformer-based time-series model; and RATD \cite{liu2024retrieval}, a retrieval-augmented diffusion model for time-series forecasting. RATD is the closest related work as it also incorporates context retrieval into the diffusion process; however, it uses retrieved sequences as conditional guidance over the diffusion process, rather than using them to construct context-dependent noise for the diffusion process, as was done in our work. 
These models were used to predict future stock returns. From these forecasts, we construct portfolios by ranking the predicted returns and equally weighting the top-50 stocks at each rebalancing date. 

\textit{End-to-end portfolio optimization models.} We also compare against the same baselines used in the original SDF work: BSV \cite{brandt2009parametric}, a linear characteristic-based model; DKKM \cite{didisheim2024apt}, a high-dimensional non-linear model based on random features; MLP, a deep neural network using own-asset characteristics; Linear Attention (LinAttn), which introduces cross-asset interactions through attention; and SDF \cite{kelly2025artificial}, a transformer-based model that incorporates cross-asset information sharing, which performed the best in the SDF work. These models directly learn portfolio weights from the input features.

We evaluate portfolio performance using returns-based, risk-based, and risk-adjusted portfolio metrics. The cumulative and annualized returns measure the overall profitability of the portfolio. However, profitability metrics alone may be insufficient, as high returns can usually be achieved by taking on excessive risk. To account for this, we include risk-based metrics such as maximum drawdown and volatility, which capture the downside risk and the variability of returns. Finally, we report risk-adjusted metrics including the Sharpe, Sortino, and Calmar ratios, which assess the efficiency of returns relative to different notions of risk, and serve as the main indicators of portfolio quality.

\paragraph{Implementation Details.} \label{para:implementation} \textsc{radar} uses GRU-based encoders with additive attention pooling for both price and text modalities, fused via a bilinear layer. The hidden dimension is set to 64. The diffusion module uses $T=100$ timesteps with a linear noise schedule. We retrieve the top $K=10$ nearest neighbors from the context bank, which is refreshed every 5 epochs. We use an input sequence length of 60 trading days and a rebalancing horizon of $H=7$ days. The model is optimized with AdamW and cosine annealing over 50 epochs with early stopping ($\text{patience}=10$). Gradients are clipped at 1.0. All experiments are conducted on a single NVIDIA GeForce RTX 4090 GPU.

%% file: 5_results.tex
\section{Results} 

\begin{table}[h]
\vspace{-10px}
\centering
\caption{Performance comparison. The best baselines are underlined, and the best results are bolded.}
\resizebox{1.0\textwidth}{!}{
\begin{tabular}{lrrrrrrr}
\toprule
Model & Sharpe (\(\uparrow\)) & Sortino (\(\uparrow\)) & Calmar (\(\uparrow\)) & CumRet (\(\uparrow\)) & AnnRet (\(\uparrow\)) & MaxDD (\(\downarrow\)) & Vol (\(\downarrow\)) \\
\midrule
\multicolumn{8}{l}{\textbf{Benchmarks}} \\
S\&P 500 & 0.513 & 0.576 & 0.244 & 0.487 & 0.083 & 0.339 & 0.191 \\
Equal Weight \cite{demiguel2009optimal} & \underline{0.795} & 0.873 & 0.393 & 1.005 & 0.149 & 0.380 & 0.200 \\
\multicolumn{8}{l}{\textbf{Forecasting}} \\
HAN \cite{hu2018listening} & 0.705 & 0.816 & 0.355 & 0.892 & 0.123 & 0.347 & 0.191 \\
StockNet \cite{xu2018stock} & 0.780 & \underline{0.899} & 0.387 & \underline{1.218} & \underline{0.156} & 0.404 & 0.216 \\
iTransformer \cite{liu2023itransformer} & 0.758 & 0.877 & 0.393 & 0.997 & 0.134 & 0.342 & 0.190 \\
NGAT \cite{niu2025ngat} & 0.710 & 0.817 & 0.334 & 0.893 & 0.123 & 0.369 & 0.189 \\
RATD \cite{liu2024retrieval} & 0.680 & 0.754 & 0.302 & 0.844 & 0.130 & 0.431 & 0.214 \\
\multicolumn{8}{l}{\textbf{Portfolio Opt}} \\
BSV \cite{brandt2009parametric} & 0.361 & 0.481 & 0.164 & 0.252 & 0.046 & 0.281 & 0.160 \\
DKKM \cite{didisheim2024apt} & 0.367 & 0.489 & 0.177 & 0.229 & 0.042 & 0.238 & 0.139 \\
LinAttn & 0.407 & 0.558 & 0.326 & 0.195 & 0.036 & \underline{0.111} & 0.100 \\
MLP & 0.502 & 0.710 & \underline{0.406} & 0.203 & 0.038 & \textbf{0.093} & \textbf{0.080} \\
SDF \cite{kelly2025artificial} & 0.505 & 0.690 & 0.313 & 0.211 & 0.039 & 0.125 & \underline{0.083} \\
\multicolumn{8}{l}{\textbf{Our Model}} \\
\textsc{radar} 
& \textbf{1.062} & \textbf{1.352} & \textbf{0.924} & \textbf{2.502} & \textbf{0.285} & 0.308 & 0.271 \\
% & \textbf{0.994} & \textbf{1.276} & \textbf{0.720} & \textbf{2.502} & \textbf{0.285} & 0.308 & 0.271 \\
\bottomrule
\end{tabular}
}
\vspace{-5px}
\label{tab:main_results}
\end{table}

\paragraph{Performance Comparison.} 
Table \ref{tab:main_results} reports the forecasting results. We can observe the following:
\begin{itemize}[leftmargin=*]
\item The Equal Weight portfolio obtained strong results across all metrics, and achieved the second-highest Sharpe ratio. Prior work \cite{demiguel2009optimal} have shown that it is often difficult to outperform naive equal-weighted diversification in practice, as forecasted returns are often noisy, and portfolio optimization strategies amplifies any forecasting errors. Equal weighting avoids this issue by ignoring forecasts altogether and therefore remains highly robust on the out-of-sample test data.

\item The forecasting models achieved close performance, with StockNet delivering the second-best results on some metrics. Here, the portfolio is constructed by buying the forecasted top-50 assets, which allows these models to perform better on the returns compared to the portfolio methods, but also realizing higher volatilities. However, we note that their results remain relatively consistent with each other and similar to equal-weighting regardless of model forecasting performance. By buying a larger number of assets, any forecasting errors for an asset get neutralized by the others, showing that forecasting ability might not matter much in a real-life multi-asset portfolio setting.

\item The portfolio methods show results that are largely consistent with those reported in the original SDF work \cite{kelly2025artificial}. The Sharpe performance increases from BSV which is linear, across higher levels of model expressiveness in DKKM, LinAttn and MLP. The original SDF model introduces cross-asset information sharing, which outperforms the others. In that sense, our results are consistent with the reported takeaway that both model expressiveness and cross-asset structures matter for portfolio performance. We note that these models demonstrate lower drawdowns and volatility than the forecasting models due to not optimizing only for returns. However, their Sharpe ratios remain lower, which might be due to their lower expressiveness compared to the deep-learning models.

\item \textsc{radar} attains the best performance on most metrics including the Sharpe, Sortino, and Calmar ratios, indicating the best overall risk-return trade-off among all methods. It also achieved lower drawdowns  than the forecasting models. The non-ratios metric performances could be less crucial here, as the model seeks to optimize for the best trade-offs to get the best portfolio results.
\end{itemize}

We also provide statistical significance and robustness analyses in Appendix~\ref{app:significance}. 
% Additionally, more detailed analyses such as the cross-sectional performances and financial factor studies (including the portfolio alpha results) are presented in Appendix~\ref{app:cross_section} and Appendix~\ref{app:financial}, respectively.

\paragraph{Ablation Study.} We conduct an ablation study to demonstrate the effectiveness of the model design.

\begin{table}[h]
\centering
\caption{Ablation study across different data and component variations of the \textsc{radar} framework.}
\resizebox{1.0\textwidth}{!}{
\begin{tabular}{lrrrrrrr}
\toprule
Model & Sharpe (\(\uparrow\)) & Sortino (\(\uparrow\)) & Calmar (\(\uparrow\)) & CumRet (\(\uparrow\)) & AnnRet (\(\uparrow\)) & MaxDD (\(\downarrow\)) & Vol (\(\downarrow\)) \\
\midrule

\multicolumn{8}{l}{\textbf{Data}} \\
w/o News & 0.650 & 0.821 & 0.304 & \underline{1.229} & \underline{0.174} & 0.573 & 0.332 \\
w/o Price & 0.797 & 0.878 & 0.395 & 1.003 & 0.149 & 0.377 & \underline{0.200} \\

\multicolumn{8}{l}{\textbf{Components}} \\
SDF (baseline) & 0.505 & 0.690 & 0.313 & 0.211 & 0.039 & \textbf{0.125} & \textbf{0.083} \\
\textit{+ Embeddings} & \underline{0.859} & \underline{0.959} & \underline{0.429} & 1.138 & 0.164 & 0.383 & 0.201 \\
\textit{+ Diffusion} & 0.411 & 0.466 & 0.108 & 0.501 & 0.085 & 0.784 & 0.525 \\

\multicolumn{8}{l}{\textbf{Our Model}} \\
\textsc{radar} 
& \textbf{1.062} & \textbf{1.352} & \textbf{0.924} & \textbf{2.502} & \textbf{0.285} & \underline{0.308} & 0.271 \\
% & \textbf{0.994} & \textbf{1.276} & \textbf{0.720} & \textbf{2.502} & \textbf{0.285} & 0.308 & 0.271 \\

\bottomrule
\end{tabular}
}
\label{tab:ablation_results}
\end{table}

\begin{itemize}[leftmargin=*]
\item From Table \ref{tab:ablation_results}, removing either of the news or price inputs degrade performance, showing that both information sources are crucial for model performance. Among the two, the No Price variant shows relatively low volatility results, which suggests that much of the volatility is driven by the noisy price data. This is consistent with prior literature \cite{demiguel2009optimal, koa2023diffusion} which show that price information often contain noise, leading to more volatile forecasts, which are further amplified in portfolio strategies.

\item Moving from the SDF baseline to \textit{+Embeddings} yields a clear improvement across most metrics. In the original SDF work \cite{kelly2025artificial}, the portfolio learning component operates directly on the raw features, whereas we first learn a market representation before optimization. The improvement suggests that the representation learning helps to organize the market data into a cleaner latent space, separating the information from noise and making the downstream portfolio optimization more effective.

\item Moving from \textit{+Embeddings} to \textit{+Diffusion} unexpectedly reduces performance. A plausible explanation is that the learnt embeddings already capture most of the useful information available in the data, so further corrupting them with unrelated Gaussian noise weakens the representation rather than improving it. This could be corroborated by the degraded performance in the max drawdown and volatility metrics. Here, diffusion noise appears to act more as distortion than regularization.

\item \textsc{radar} achieves the best performance across most key metrics. Unlike \textit{+Diffusion}, \textsc{radar} introduces a context bank that retrieves relevant historical states, hence the added noise is no longer arbitrary but grounded in meaningful past market conditions. This provides a set of plausible future states over which the portfolio can be optimized, resulting in a more robust risk-return trade-off.
\end{itemize}

\paragraph{Parameter Selection.} 
For the context bank, we do a study over different values of top $K$, which determines the number of relevant historical segments used to determine the diffusion parameters.

\begin{figure}[h]
    \centering
    \begin{subfigure}[t]{0.45\columnwidth}
        \centering
        \includegraphics[width=\linewidth]{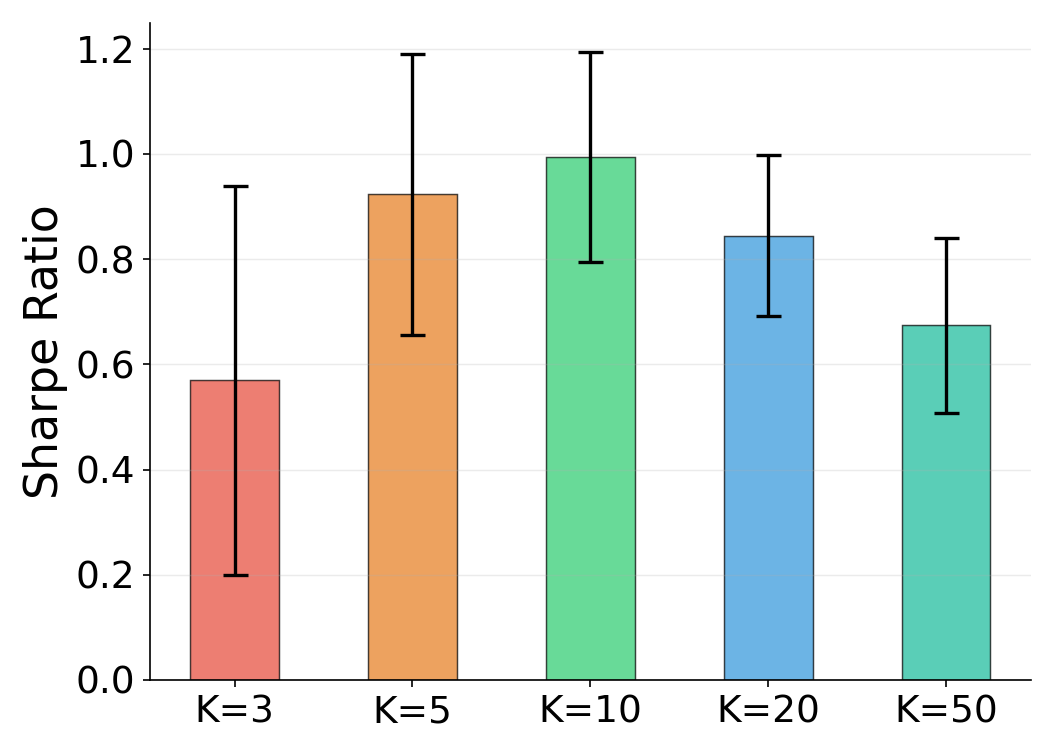}
        \caption{Sharpe Ratio (mean $\pm$ std)}
        \label{fig:sharpe_bar}
    \end{subfigure}
    \hfill
    \begin{subfigure}[t]{0.45\columnwidth}
        \centering
        \includegraphics[width=\linewidth]{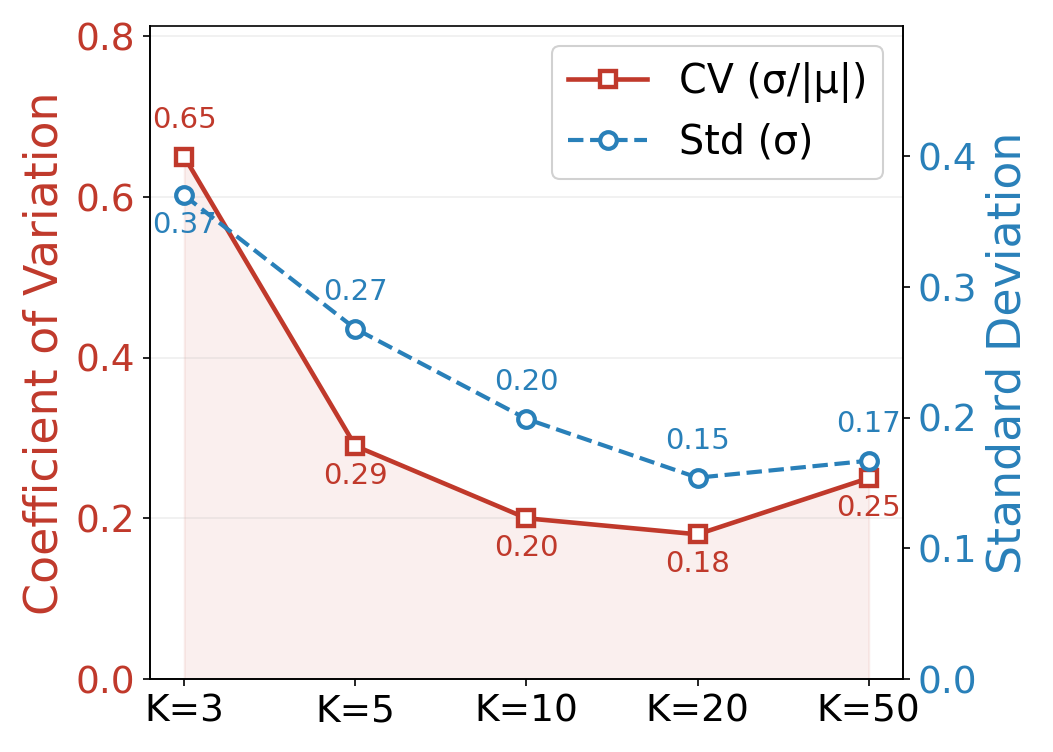}
        \caption{Cross-Seed Instability: CV and Std}
        \label{fig:cv_trend}
    \end{subfigure}
    \caption{Effect of different top-$K$ values on performance and stability across random seeds.}
    \label{fig:topk_stability}
\end{figure}

From Figure \ref{fig:topk_stability}, we see a clear trade-off between retrieval robustness and signal fidelity as $K$ varies. When $K$ is small, performance exhibits substantial cross-seed variability, indicating that the retrieval process is not stable. Intuitively, relying on only a few retrieved neighbors makes the estimated signal highly sensitive to small perturbations, which propagates into downstream portfolio outcomes.

On the other hand, at larger values of $K$, although variability remains low, the mean Sharpe performance begins to deteriorate. This indicates that incorporating too many retrieved samples also introduces less relevant information, effectively diluting the signal and reducing overall effectiveness.

We chose $K=10$ as our setting, which shows strong performance while maintaining low variability.

\paragraph{Market Representation Study.} 
Our work aims to learn market representations that improve portfolio outcomes. We study whether the denoised embeddings capture meaningful market structure.

\begin{wrapfigure}{r}{0.6\textwidth}
    \centering
    \includegraphics[width=0.6\textwidth]{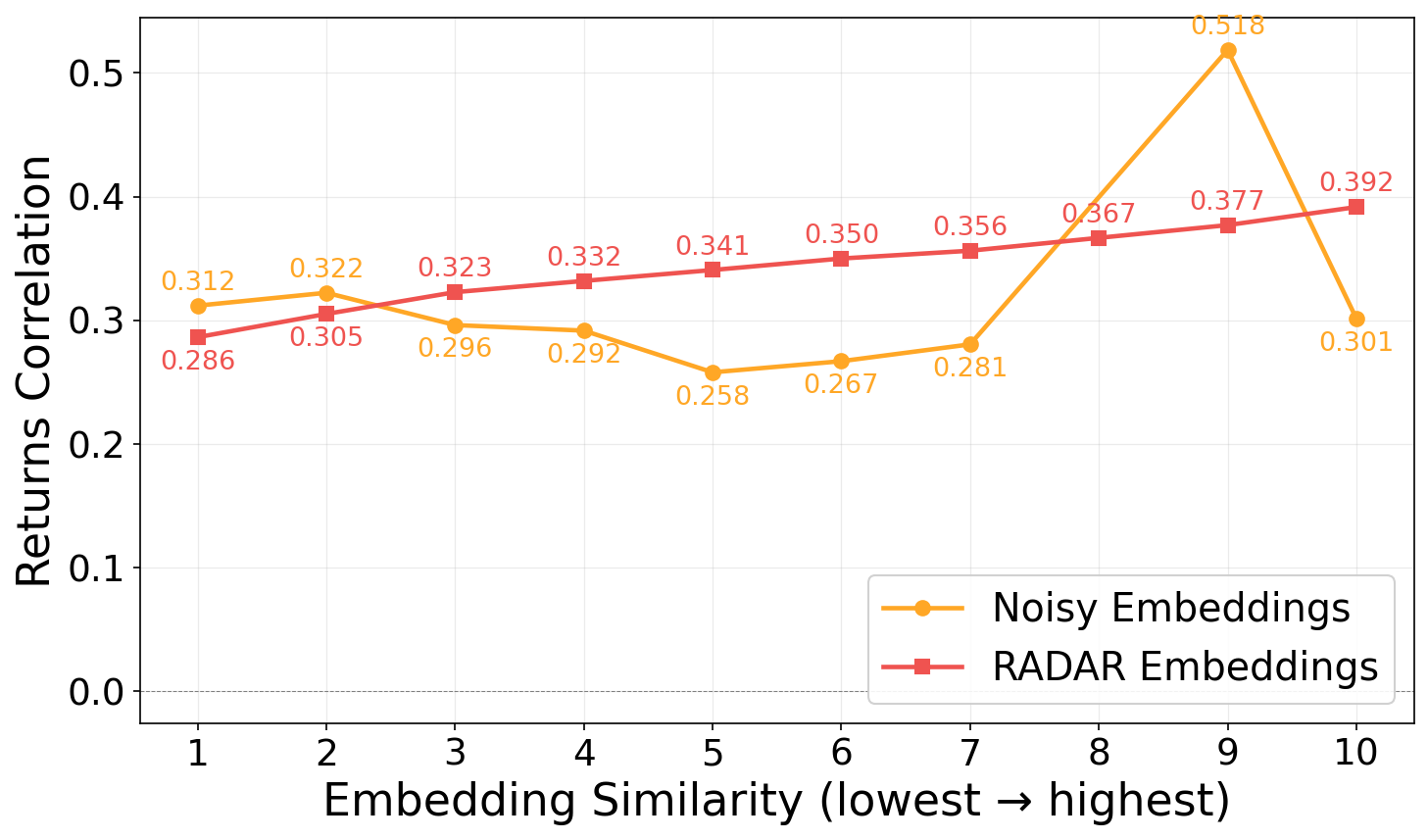}
    \caption{Embedding Similarity \textit{vs} Returns Correlation. Reported values are aggregated over 5 rolling windows.}
    \vspace{-20pt}
    \label{fig:sim-vs-corr}
\end{wrapfigure}

In Figure \ref{fig:sim-vs-corr}, we study the asset correlation information (required for portfolio optimization) captured by the original noisy embeddings $\mathbf{m}_t$ and the cleaned \textsc{radar} embeddings $\hat{\mathbf{z}}_t$. 

Here, all asset pairs are ranked by cosine similarity in the embedding space and grouped into deciles, from least (Decile 1) to most similar (Decile 10).
For each decile, we compute the average realised return correlation between the corresponding asset pairs.

A meaningful representation should produce a monotonic relationship: higher embedding similarities should correspond to stronger correlations.

As shown, the \textsc{radar} embeddings yield a clearer and more monotonic increase in correlation across deciles compared to the noisy embeddings.
This indicates that the context-aware denoising process helps to improve alignment between embedding similarity and true market co-movement.

\begin{wrapfigure}{r}{0.6\textwidth}
    \centering
    \includegraphics[width=0.6\textwidth]{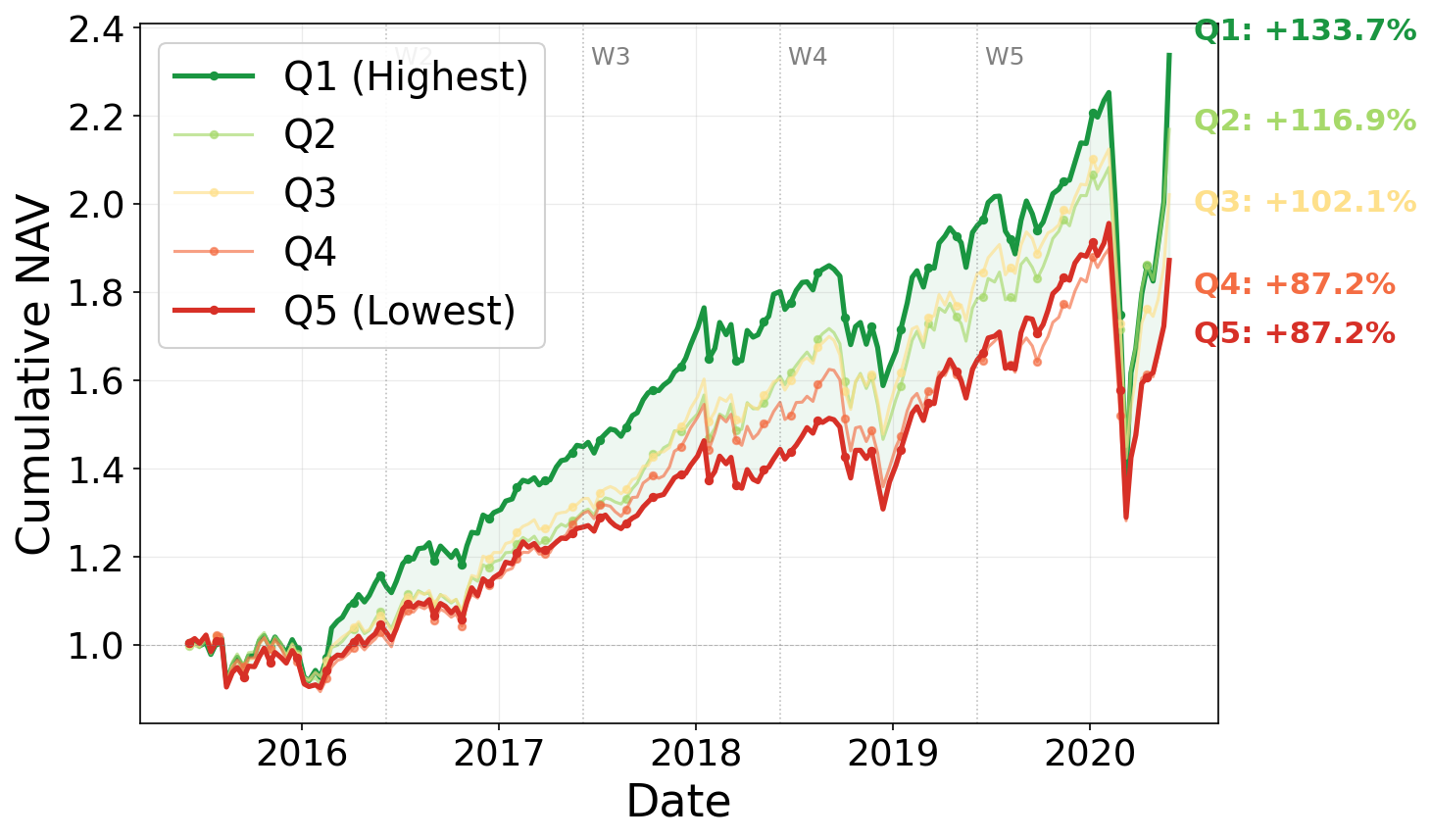}
    \caption{Quintile Portfolio NAV by Asset Scoring.}
    \label{fig:quintile-nav}
    \vspace{-10pt}
\end{wrapfigure}

In Figure \ref{fig:quintile-nav}, we evaluate whether the cleaned \textsc{radar} embeddings $\hat{\mathbf{z}}_t$ capture any useful information about the constituent asset returns performance, which are also required for portfolio optimization.

At each rebalancing date, the learnt asset embeddings are mapped to scalar scores via a MLP scoring network.
Assets are then ranked by these scores and partitioned into quintiles, from highest (Q1) to lowest (Q5).
Each quintile forms an equal-weighted portfolio, and we track cumulative NAV over time using forward returns.

A meaningful representation should induce a monotonic ordering: portfolios formed from the higher-scoring assets should outperform those formed from lower-scoring assets.

As shown, the NAV curves exhibit a clear separation, with Q1 consistently outperforming the lower quintiles.
This indicates that the \textsc{radar} embeddings capture economically meaningful cross-sectional asset returns signals that translate into improved portfolio performance.

% From these results, we find that the \textsc{radar} embeddings encode useful information on both the asset returns and correlations, which are both crucial components for the portfolio optimization task.

\begin{figure}[h]
    \centering
    \includegraphics[width=\textwidth]{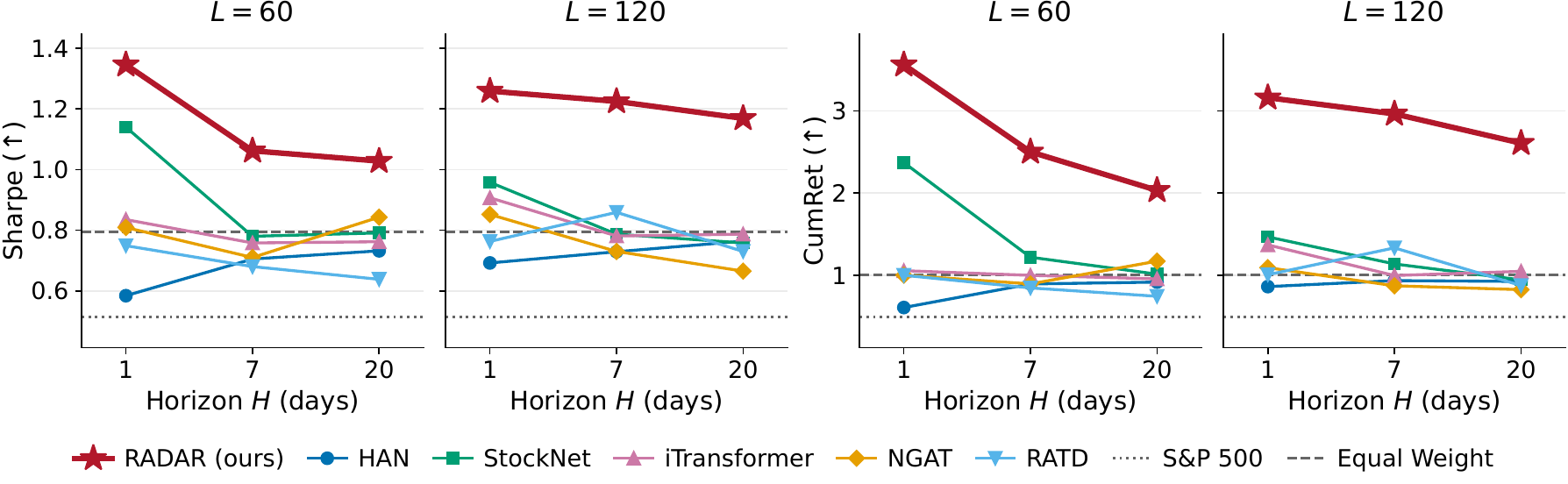}
    \caption{Performance across different input length $L$ and rebalancing horizon $H$. Each pair shares its $y$-axis. S\&P~500 and Equal Weight do not depend on $L$ and $H$, and are drawn as reference lines.}
    \vspace{-7px}
    \label{fig:length_analysis}
\end{figure}

\paragraph{Robustness to Context and Target Lengths.}
We further evaluate the robustness of \textsc{radar} across different input and output lengths by varying the input length $L\in\{60,120\}$ and rebalancing horizon $H\in\{1,7,20\}$. As shown in Figure \ref{fig:length_analysis}, \textsc{radar} consistently achieves the highest Sharpe ratio and cumulative return across all six configurations. Performance generally decreases as the rebalancing horizon increases as longer horizons introduce greater uncertainty in future price movements, but \textsc{radar} remains consistently above the forecasting baselines. These results demonstrate that the performance gains of \textsc{radar} are not specific to any particular choice of input or output lengths.

\paragraph{Generalizability to Other Domains.}
For generalizability, we further evaluate \textsc{radar} on Time-MMD~\cite{liu2024time}, a multi-modal (text + numerical) time-series benchmark spanning multiple domains.

\begin{wraptable}{r}{0.50\linewidth}
    \vspace{-1.0\baselineskip}
    \centering
    \small
    \caption{Generalizability on different domains.}
    \vspace{-7px}
    \label{tab:timemmd}
    \begin{tabular}{lccc}
        \toprule
        \textbf{Domain} & \textbf{No Diff.} & \textbf{Gaussian} & \textbf{\textsc{radar}} \\
        \midrule
        Agriculture  & \underline{2.328} & 2.375 & \textbf{2.240} \\
        Climate      & 0.484 & \textbf{0.464} & \textbf{0.464} \\
        Energy       & 1.753 & \underline{0.540} & \textbf{0.517} \\
        Environment  & \underline{0.931} & 0.946 & \textbf{0.928} \\
        Health (AFR) & \underline{2.413} & 4.203 & \textbf{1.635} \\
        Health (US)  & 1.007 & \underline{0.956} & \textbf{0.821} \\
        Security     & \textbf{1.526} & 1.584 & \underline{1.557} \\
        SocialGood   & 0.953 & \underline{0.547} & \textbf{0.512} \\
        Traffic      & \textbf{1.154} & 1.350 & \underline{1.290} \\
        \bottomrule
    \end{tabular}
    \vspace{-1.0\baselineskip}
\end{wraptable}

Overall, we find that \textsc{radar} is largely generalizable across multiple domains. In some cases such as Security and Traffic, \textit{No Diffusion} performs the best, which might be attributed to the lack of noise in the data. In other cases, injecting \textit{Gaussian} noise does not appear to degrade performance, unlike the observations in our financial domain. This might indicate that the noise distribution of data in these domains is closer to Gaussian. These observations could be studied further in future work, based on the specific noise characteristics of each domain. 

% \paragraph{Additional Results.}
% More detailed performance analyses on the \textsc{radar} framework are provided in the appendix. 
% In particular, Appendix~\ref{app:cross_section} presents cross-sectional performance across different volatility groups and industry sectors, highlighting how performance varies under different market conditions. Appendix~\ref{app:financial} reports financial factor analyses, including CAPM and Fama--French regressions, along with portfolio alpha studies to further assess whether returns can be explained by the standard risk factors.

\paragraph{Additional Results.}
More detailed performance analyses on the \textsc{radar} framework are provided in the appendix. In particular, Appendix~\ref{app:cross_section} reports cross-sectional analyses across volatility groups and industry sectors, while Appendix~\ref{app:financial} provides financial factor analyses and portfolio alpha studies.

%% file: 6_conclusion.tex
\section{Conclusion} \label{sec:conclusion}
In this work, we studied the portfolio task under the stochastic discount factor (SDF) framework. We explore learning clean market representations to better capture the underlying structure of financial data to perform portfolio optimization. This problem is challenging due to several factors: financial markets exhibit time-varying dynamics with shifting regimes, the multimodal inputs of price and news data often contain noise, and isotropic noise do not capture the state-dependent and heteroskedastic nature of financial uncertainty. To address these issues, we introduced \textsc{radar}, a retrieval-augmented diffusion framework that learns context-aware market representations by conditioning on similar historical regimes to denoise multimodal features. We conducted extensive experiments on forecasting and portfolio models, and showed that \textsc{radar} achieves strong performance on key risk-adjusted metrics, while producing representations that align with assets co-movement and returns structure. 

\paragraph{Limitations.} \label{para:limitations} Our experiments rely on the FNSPID news dataset \cite{dong2024fnspid}, which is relatively sparse in coverage. Since our retrieval mechanism depends on the availability of relevant historical contexts, limited news coverage would affect the quality of the learnt representations. However, we also note that the FNSPID dataset is widely used in literature \cite{liu2025fin, liu2024time, agarwal2024prompt}, whereas more comprehensive but proprietary sources (\eg Bloomberg, Reuters) could reduce the overall replicability of the work. 

Compared to classical SDF approaches \cite{kelly2024large, kelly2025artificial} that operate directly on raw characteristics, our use of embeddings to optimize portfolio weights may reduce interpretability, as the resulting representations are less directly tied to economically meaningful factor inputs. This limits the ability of the framework to capture feature importance. However, this limitation could be mitigated by applying post-hoc explanation techniques such as SHAP \cite{lundberg2017unified} or LIME \cite{ribeiro2016should}, which we will leave for future work.

%% file: a_proof_final.tex
\section{Diffusion Process Formulation}
\label{app:diffusion_formulation}

The standard diffusion process corrupts data with Gaussian noise of
$\mathcal{N}(\mathbf{0},\mathbf{I})$. In Equations \eqref{eq:forward_q} and \eqref{eq:forward_c}, we instead corrupt the representations with Gaussian noise parameterized by
$(\boldsymbol{\mu}_{\mathcal{N}_{q,t}},
\boldsymbol{\sigma}_{\mathcal{N}_{q,t}})$, which are the the retrieved statistics.
A shifted, rescaled Gaussian remains Gaussian, so the process is still a standard diffusion process expressed in new coordinates. We can derive the formulation:

\paragraph{Forward Process.}
Equation \eqref{eq:forward_q} gives the marginal distribution at an arbitrary diffusion step $s$:
\begin{equation*}
\tilde{\mathbf{q}}_{t,s}
=
\sqrt{\bar{\alpha}_s}\,\mathbf{q}_t
+
\sqrt{1-\bar{\alpha}_s}
\left(
\boldsymbol{\mu}_{\mathcal{N}_{q,t}}
+
\boldsymbol{\sigma}_{\mathcal{N}_{q,t}}
\odot
\boldsymbol{\epsilon}
\right).
\end{equation*}

For compactness, we define:
\begin{equation*}
r_s = \sqrt{1-\bar{\alpha}_s},
\qquad
\boldsymbol{\Sigma}_{q,t}
=
\operatorname{diag}
\left(
\boldsymbol{\sigma}_{\mathcal{N}_{q,t}}^2
\right),
\end{equation*}

and let $\alpha_s = 1-\beta_s$. A stepwise forward process consistent with Equation \eqref{eq:forward_q} is:
\begin{equation*}
q\left(
\tilde{\mathbf{q}}_{t,s}
\mid
\tilde{\mathbf{q}}_{t,s-1},
\mathcal{N}_{q,t}
\right)
=
\mathcal{N}
\left(
\sqrt{\alpha_s}\,
\tilde{\mathbf{q}}_{t,s-1}
+
\delta_s
\boldsymbol{\mu}_{\mathcal{N}_{q,t}},
\;
\beta_s\boldsymbol{\Sigma}_{q,t}
\right),
\end{equation*}
where:
\begin{equation*}
\delta_s
=
r_s-\sqrt{\alpha_s}\,r_{s-1}.
\end{equation*}

If we define a mean-centered variable:
\begin{equation*}
\mathbf{y}^{q}_{t,s}
=
\tilde{\mathbf{q}}_{t,s}
-
r_s\boldsymbol{\mu}_{\mathcal{N}_{q,t}},
\end{equation*}
then:
\begin{equation*}
\mathbf{y}^{q}_{t,s}
=
\sqrt{\alpha_s}\,
\mathbf{y}^{q}_{t,s-1}
+
\sqrt{\beta_s}\,
\boldsymbol{\sigma}_{\mathcal{N}_{q,t}}
\odot
\boldsymbol{\epsilon}_s.
\end{equation*}

This is the standard diffusion formulation, after subtracting the context-dependent mean and scaling the noise by the context-dependent standard deviation.

\paragraph{Reverse Posterior.}
Given the clean representation $\mathbf{q}_t$, the exact posterior is:
\begin{equation*}
q\left(
\tilde{\mathbf{q}}_{t,s-1}
\mid
\tilde{\mathbf{q}}_{t,s},
\mathbf{q}_t,
\mathcal{N}_{q,t}
\right)
=
\mathcal{N}
\left(
\tilde{\boldsymbol{\mu}}_{q,s},
\;
\tilde{\beta}_s
\boldsymbol{\Sigma}_{q,t}
\right),
\end{equation*}
where:
\begin{equation*}
\tilde{\beta}_s
=
\frac{
\beta_s(1-\bar{\alpha}_{s-1})
}{
1-\bar{\alpha}_s
},
\end{equation*}
and:
\begin{equation*}
\tilde{\boldsymbol{\mu}}_{q,s}
=
A_s\mathbf{q}_t
+
B_s
\left(
\tilde{\mathbf{q}}_{t,s}
-
r_s\boldsymbol{\mu}_{\mathcal{N}_{q,t}}
\right)
+
r_{s-1}\boldsymbol{\mu}_{\mathcal{N}_{q,t}},
\end{equation*}
with:
\begin{equation*}
A_s
=
\frac{
\beta_s\sqrt{\bar{\alpha}_{s-1}}
}{
1-\bar{\alpha}_s
},
\qquad
B_s
=
\frac{
\sqrt{\alpha_s}(1-\bar{\alpha}_{s-1})
}{
1-\bar{\alpha}_s
}.
\end{equation*}

\paragraph{Link to Equation \eqref{eq:denoising}.}
The true $\mathbf{q}_t$ is unavailable during reverse sampling. The denoising function predicts it as:
\begin{equation*}
\hat{\mathbf{q}}_t
=
D_p
\left(
\tilde{\mathbf{q}}_{t,s},
s
\mid
\mathbf{m}_t
\right).
\end{equation*}

We define the learned reverse process by replacing the unknown $\mathbf{q}_t$ in the exact posterior with $\hat{\mathbf{q}}_t$:
\begin{equation*}
p_\theta
\left(
\tilde{\mathbf{q}}_{t,s-1}
\mid
\tilde{\mathbf{q}}_{t,s},
\mathbf{m}_t,
\mathcal{N}_{q,t}
\right)
=
\mathcal{N}
\left(
\boldsymbol{\mu}_{\theta,q,s},
\;
\tilde{\beta}_s
\boldsymbol{\Sigma}_{q,t}
\right),
\end{equation*}
where:
\begin{equation*}
\boldsymbol{\mu}_{\theta,q,s}
=
A_s
D_p
\left(
\tilde{\mathbf{q}}_{t,s},
s
\mid
\mathbf{m}_t
\right)
+
B_s
\left(
\tilde{\mathbf{q}}_{t,s}
-
r_s\boldsymbol{\mu}_{\mathcal{N}_{q,t}}
\right)
+
r_{s-1}
\boldsymbol{\mu}_{\mathcal{N}_{q,t}}.
\end{equation*}

Sampling one reverse step is then:
\begin{equation*}
\tilde{\mathbf{q}}_{t,s-1}
=
\boldsymbol{\mu}_{\theta,q,s}
+
\sqrt{\tilde{\beta}_s}\,
\boldsymbol{\sigma}_{\mathcal{N}_{q,t}}
\odot
\mathbf{z},
\qquad
\mathbf{z}
\sim
\mathcal{N}(\mathbf{0},\mathbf{I}).
\end{equation*}

At the final step, $s=1$,
\begin{equation*}
\tilde{\mathbf{q}}_{t,0}
=
D_p
\left(
\tilde{\mathbf{q}}_{t,1},
1
\mid
\mathbf{m}_t
\right).
\end{equation*}

The same formulation applies to the text path.

\paragraph{Deriving Equation \eqref{eq:diff_loss}.}
The denoising network is defined as a $x_0$-prediction model:
\begin{equation*}
D_p
\left(
\tilde{\mathbf{q}}_{t,s},
s
\mid
\mathbf{m}_t
\right)
\approx
\mathbf{q}_t.
\end{equation*}
Under the reverse formulation above, the exact variational term for $s>1$ is proportional to:
\begin{equation*}
\lambda_s
\left\|
\boldsymbol{\Sigma}_{q,t}^{-1/2}
\left(
\mathbf{q}_t-\hat{\mathbf{q}}_t
\right)
\right\|_2^2,
\end{equation*}
where:
\begin{equation*}
\lambda_s
=
\frac{A_s^2}{2\tilde{\beta}_s}.
\end{equation*}

Equation \eqref{eq:diff_loss} is the simplified $x_0$-prediction objective obtained by dropping the timestep-dependent and covariance-dependent weighting $\lambda_s$ and $\boldsymbol{\Sigma}_{q,t}^{-1/2}$:
\begin{equation*}
\mathcal{L}_{\mathrm{Diff}}
=
\mathbb{E}_{t,s}
\left[
\left\|
\mathbf{q}_t-\hat{\mathbf{q}}_t
\right\|_2^2
+
\left\|
\mathbf{c}_t-\hat{\mathbf{c}}_t
\right\|_2^2
\right].
\end{equation*}
This follows established diffusion practices: the classic denoising diffusion probabilistic model \cite{ho2020denoising} discards the timestep-dependent variational weighting in its objective, while other works \cite{hoogeboom2023blurring} also adopt an unweighted squared-error objective for a generalized non-isotropic diffusion process.

%% file: b_appendix.tex
% \appendix
% \AppendixTOC

\section{Statistical Significance and Robustness}
\label{app:significance}

We evaluate if the performance improvements of \textsc{radar} over baselines are statistically significant.

\paragraph{Difference series.}
For each baseline, we construct the out-of-sample returns difference series:
\begin{equation*}
\text{Difference} = r_t^{\text{\textsc{radar}}} - r_t^{\text{baseline}},
\end{equation*}
and test the one-sided hypothesis that \textsc{radar} portfolio returns outperforms the baselines.

\paragraph{Pooled time-series tests.}
We first perform significance tests on the pooled return differences:

\begin{itemize}[leftmargin=*]
\item \textbf{Monthly $t$-test:} Aggregates daily returns to the monthly level before testing. This reduces the microstructure noise and short-term dependence, providing a lower-frequency validation to show that the performance gains persist beyond daily fluctuations and transient market effects.

\item \textbf{Newey--West HAC $t$-test} \cite{newey1986simple}: Adjusts for autocorrelation and heteroskedasticity in the return difference series. This is important because daily financial returns violate the i.i.d. assumption, and naive $t$-tests would otherwise systematically overstate statistical significance.

\item \textbf{Diebold--Mariano (DM) test} \cite{diebold2002comparing}: Evaluates differences in predictive accuracy over time by comparing the loss (return differences) sequence directly. Unlike standard $t$-tests, it is specifically designed for time-series forecast comparisons and remains statistically valid under serial dependence.

\item \textbf{Ledoit--Wolf (LW) test} \cite{ledoit2008robust}: Tests for differences in Sharpe ratios while accounting for estimation error in the mean and covariance of returns. This is particularly important because Sharpe ratios are non-linear functionals of returns and cannot be reliably compared using standard $t$-tests.
\end{itemize}

\begin{table}[h]
\centering
\small
% \caption{Pooled significance tests comparing \textsc{radar} against baselines. $\Delta$Sharpe denotes the improvement in Sharpe relative to baselines. Significance levels: $^{***}p<0.01$, $^{**}p<0.05$, $^{*}p<0.1$.}
\caption{Pooled significance tests comparing \textsc{radar} against baselines. The reported results are taken from observations across five random seeds. Significance levels: $^{***}p<0.01$, $^{**}p<0.05$, $^{*}p<0.1$.}
\label{tab:significance_pooled}
\begin{tabular}{lccccc}
\toprule
Baseline & Monthly $p$ & HAC $p$ & DM $p$ & LW $p$ \\
\midrule
Equal Weight & 0.0016*** & 0.0015*** & 0.0015*** & 0.0079*** \\
HAN         & 0.0015*** & 0.0016*** & 0.0016*** & 0.0105** \\
StockNet    & 0.0021*** & 0.0030*** & 0.0030*** & 0.0137** \\
iTransformer& 0.0041*** & 0.0042*** & 0.0041*** & 0.0402** \\
NGAT        & 0.0011*** & 0.0013*** & 0.0012*** & 0.0081*** \\
RATD        & 0.0003*** & 0.0005*** & 0.0005*** & 0.0018*** \\
BSV         & 0.0026*** & 0.0050*** & 0.0050*** & 0.0086*** \\
DKKM        & 0.0017*** & 0.0014*** & 0.0014*** & 0.0071*** \\
MLP         & 0.0029*** & 0.0024*** & 0.0024*** & 0.0140** \\
LinAttn     & 0.0045*** & 0.0052*** & 0.0052*** & 0.0253** \\
SDF         & 0.0046*** & 0.0036*** & 0.0035*** & 0.0189** \\
\bottomrule
\end{tabular}
\end{table}

\paragraph{Robustness checks.}
To further validate the reliability of the results, we perform robustness checks:

\begin{itemize}[leftmargin=*]
\item \textbf{Block bootstrap} \cite{politis1994stationary}: Resamples the return series using block-based sampling to preserve temporal dependence, providing a non-parametric validation that does not rely on asymptotic assumptions.

\item \textbf{Bootstrap Sharpe test:} Evaluates the significance of Sharpe ratio improvements under resampling. This complements parametric tests by directly assessing the variability of the performance metric.

\item \textbf{Fisher aggregation} \cite{fisher1934statistical}: Combines $p$-values across seeds, verifying that significance holds consistently across different random initializations rather than being driven by a subset of runs.
\end{itemize}

\begin{table}[h]
\centering
\small
\caption{Robustness statistical significance. Significance levels: $^{***}p<0.01$, $^{**}p<0.05$, $^{*}p<0.1$.}
\label{tab:significance_robust}
\begin{tabular}{lcccc}
\toprule
Baseline & Bootstrap $p$ & Boot $\Delta$Sharpe $p$ & Fisher HAC $p$ & Fisher Monthly $p$ \\
\midrule
Equal Weight & 0.0023*** & 0.0153** & 0.0012*** & 0.0020*** \\
HAN         & 0.0019*** & 0.0129** & 0.0008*** & 0.0013*** \\
StockNet    & 0.0045*** & 0.0236** & 0.0041*** & 0.0064*** \\
iTransformer& 0.0044*** & 0.0314** & 0.0024*** & 0.0036*** \\
NGAT        & 0.0020*** & 0.0126** & 0.0007*** & 0.0011*** \\
RATD        & 0.0012*** & 0.0044*** & 0.0005*** & 0.0010*** \\
BSV         & 0.0043*** & 0.0134** & 0.0086*** & 0.0139** \\
DKKM        & 0.0012*** & 0.0132** & 0.0010*** & 0.0016*** \\
MLP         & 0.0019*** & 0.0203** & 0.0007*** & 0.0011*** \\
LinAttn     & 0.0038*** & 0.0307** & 0.0048*** & 0.0072*** \\
SDF         & 0.0032*** & 0.0437** & 0.0013*** & 0.0021*** \\
\bottomrule
\end{tabular}
\end{table}

From Table \ref{tab:significance_pooled}, we observe that \textsc{radar} consistently achieves statistically significant improvements across all baselines and testing procedures. The significance holds under monthly aggregation, HAC-adjusted inference, and Sharpe-based evaluation, indicating that the results are not driven by any particular modeling assumption. From Table \ref{tab:significance_robust}, the results remain robust under block bootstrap resampling and cross-seed aggregation, indicating that the findings are stable with respect to temporal dependence and random initialization. Furthermore, the bootstrap Sharpe and Ledoit--Wolf tests confirm that the improvements in Sharpe ratio (our primary evaluation metric) are also statistically significant, which reinforces the practical relevance of the observed performance gains.

\section{Cross-Sectional Performance Analysis}
\label{app:cross_section}
We conduct experiments to analyze cross-sectional performance across volatility regimes and market sectors, providing further insights into model behavior under heterogeneous market conditions.
\begin{figure}[h]
\vspace{-5px}
\centering
\begin{subfigure}[h]{0.48\linewidth}
    \centering
    \includegraphics[width=\linewidth]{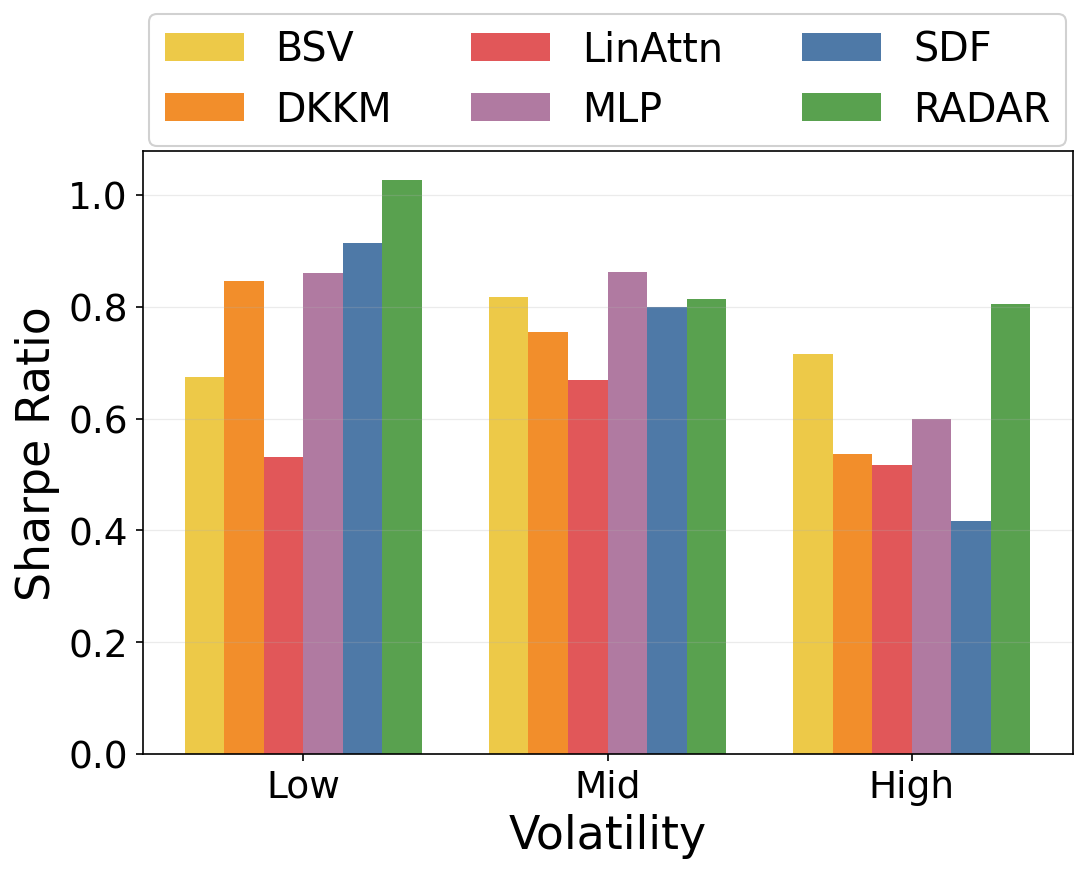}
    \caption{Sharpe Ratio across volatility groups.}
\end{subfigure}
\hfill
\begin{subfigure}[h]{0.48\linewidth}
    \centering
    \includegraphics[width=\linewidth]{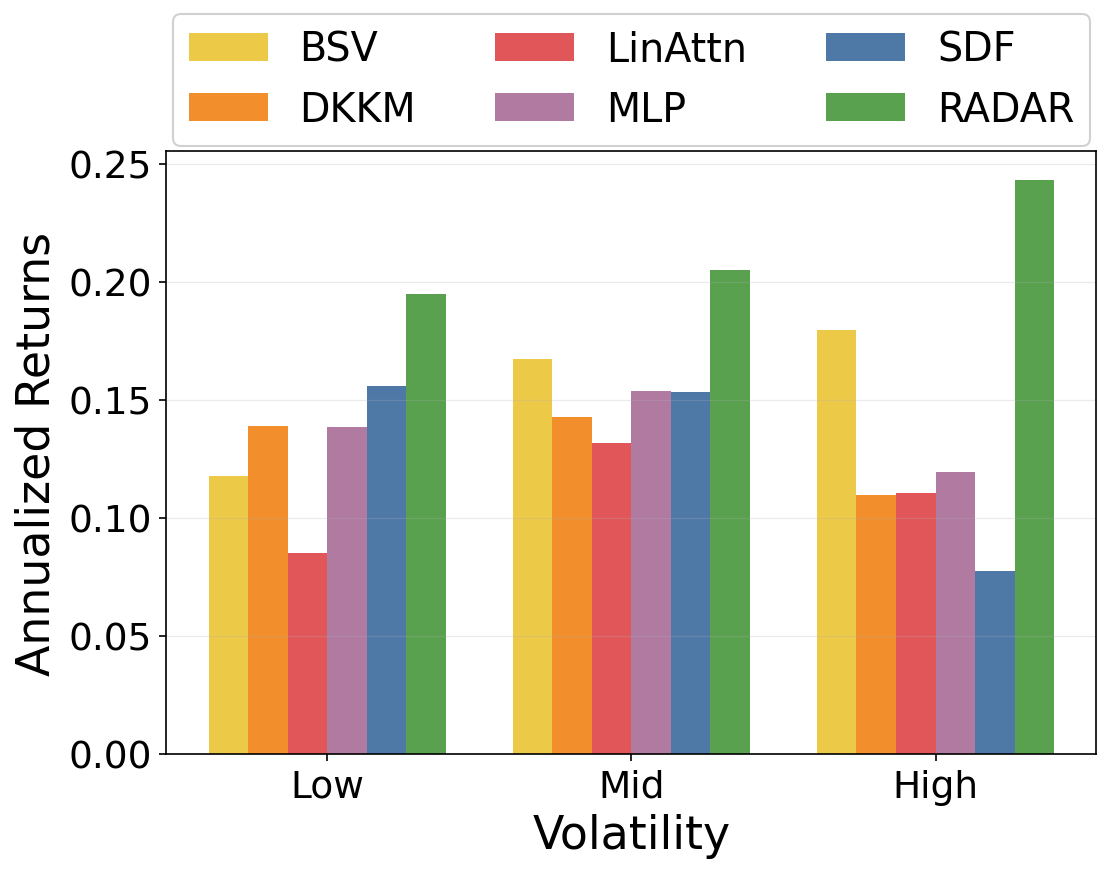}
    \caption{Annualized Returns across volatility groups.}
\end{subfigure}
\caption{Performance comparison across different volatility groups. \textsc{radar} consistently achieves higher risk-adjusted performance while maintaining competitive returns across all volatility segments.}
\label{fig:volatility_results}
\vspace{-5px}
\end{figure}

Figure \ref{fig:volatility_results} reports the results across assets sorted by volatility groups. We can observe the following:

\begin{itemize}[leftmargin=*]
\item \textsc{radar} achieves the highest annualized returns across all volatility groups, with the improvement gap widening as volatility increases. Typically, high-volatility environments are more challenging for the baseline models, as the the signal-to-noise ratio is lower, making it harder for them to extract useful information. The denoising process in \textsc{radar} helps to refine latent representations and mitigate the impact of noise, allowing the model to generalize effectively under noisy conditions.

\item The Sharpe Ratio performance trend do not strictly follow those of the annualized returns, as market volatility would cause some degradation in Sharpe (Sharpe Ratio is a ratio of returns over portfolio volatility). In terms of risk-adjusted performance, \textsc{radar} attains the highest Sharpe ratios in both the low- and high-volatility groups, and remains competitive in the mid-volatility group. This indicates that the returns improvements are not driven solely by higher risk-taking, but rather by more efficient extraction of market representative signals within each volatility regime.

\item These results are consistent with the framework design of \textsc{radar}. The retrieval-augmented component enables the model to condition on historically similar regimes, which becomes especially important in high-volatility periods where market dynamics deviate from average conditions. The diffusion-based denoising further refines the latent representations by suppressing idiosyncratic noise, allowing the model to generalize more effectively across similar contexts. Together, these mechanisms help improve stability and model performance, even in high-volatility environments.
\end{itemize}

\begin{figure}[h]
\vspace{-5px}
\centering
\includegraphics[width=0.95\linewidth]{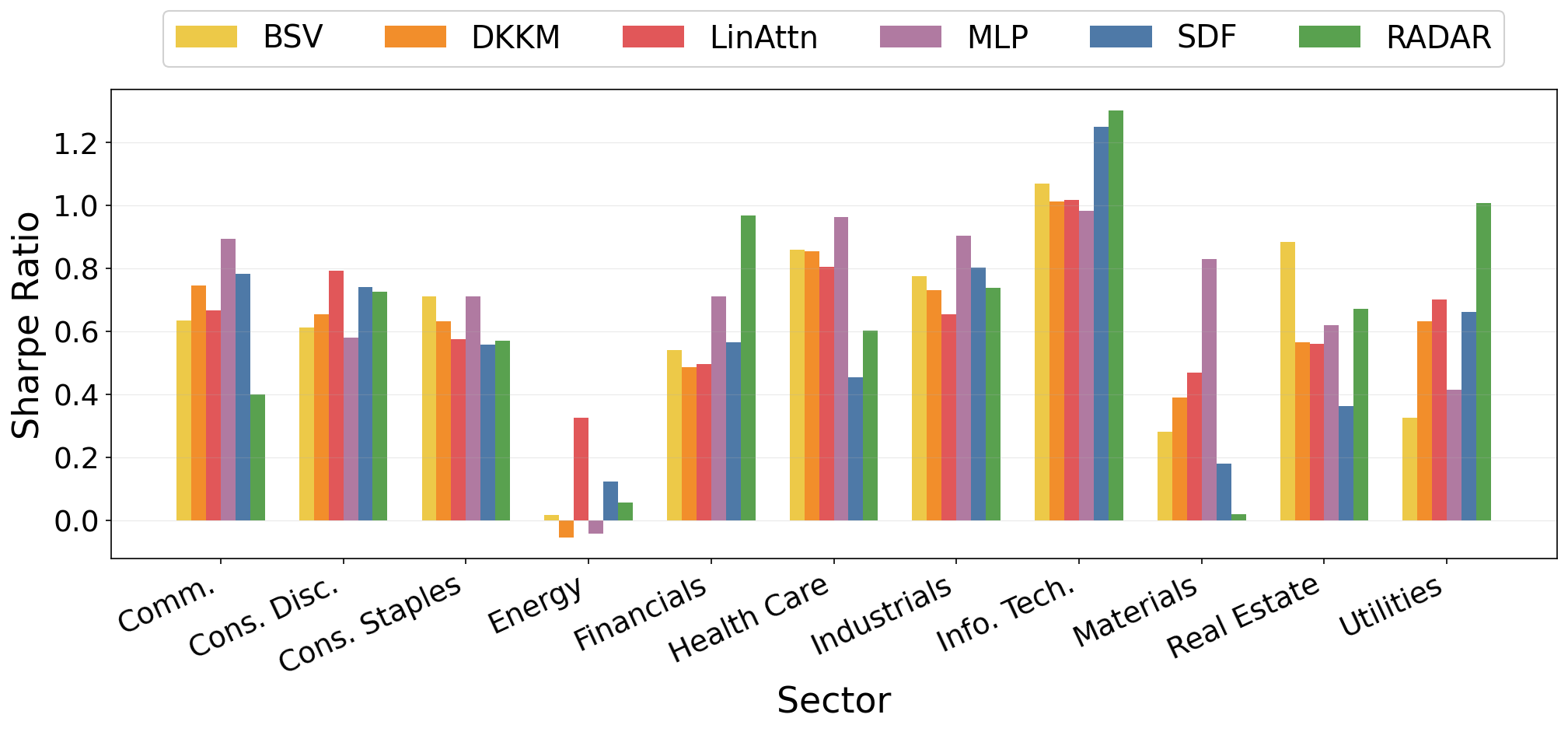}
\vspace{-5px}
\caption{Sharpe ratio across sector groups.}
\label{fig:sector_sharpe}
\vspace{-5px}
\end{figure}

\begin{figure}[h]
\centering
\includegraphics[width=0.95\linewidth]{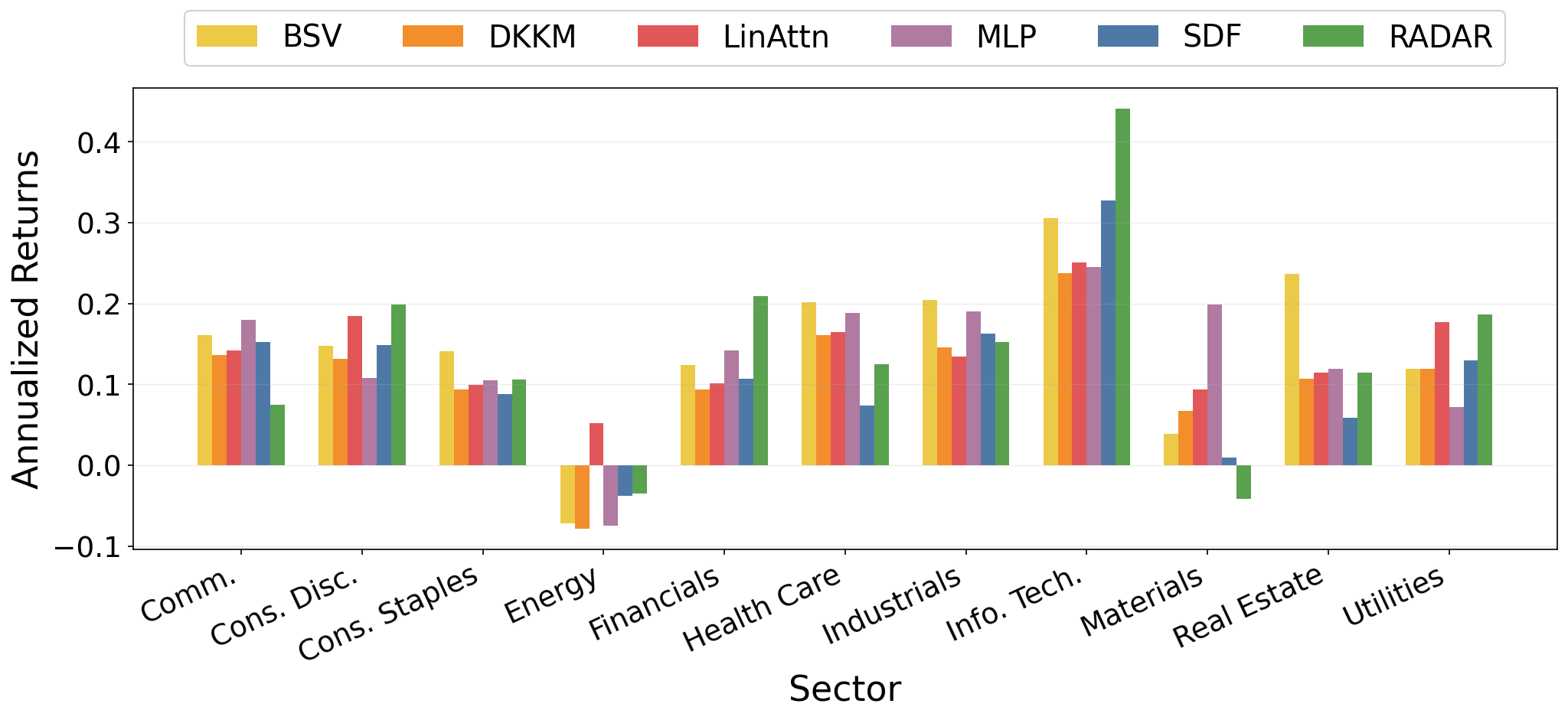}
\vspace{-5px}
\caption{Annualized returns across sector groups.}
\label{fig:sector_annret}
\vspace{-5px}
\end{figure}

Figures \ref{fig:sector_sharpe} and \ref{fig:sector_annret} present the results across assets sorted by sectors. We can observe the following:

\begin{itemize}[leftmargin=*]

\item \textsc{radar} achieves the strongest performance in most sectors, with particularly large gains in Information Technology and Financials. These sectors are typically characterized by higher cross-sectional heterogeneity and stronger interdependencies across firms. The improvements suggest that \textsc{radar} is effective at capturing complex interactions and non-linear relationships within these sectors.

\item In more stable sectors such as Utilities and Consumer Staples, \textsc{radar} remains competitive and achieves strong Sharpe performance. This suggests that the model does not rely solely on high-volatility opportunities, but is also able to extract signals within lower-variance environments.

\item We note that the gains are more limited in sectors such as Energy and Real Estate, where returns are often driven by exogenous macro factors (\eg commodity prices or interest rates) which may not be fully captured by firm-level representations that was learnt in our work. This indicates that while \textsc{radar} improves modeling of cross-sectional structure, its benefits are less pronounced in sectors dominated by external shocks. We believe that the \textsc{radar} framework could be further improved with the addition of macro-economic data, which we seek to incorporate in future work.
\end{itemize}

Overall, the results suggest that \textsc{radar} improves portfolio performance by better modeling state-dependent and noisy market environments. The retrieval-augmented mechanism enables the model to adapt to non-stationarity by conditioning on historically similar regimes, while the diffusion-based denoising refines latent representations by reducing noise in the multimodal inputs. This appears especially beneficial in high-volatility assets and sectors with complex interactions, where baselines such as MLP and transformers (SDF) show degraded performance. The initialization of the diffusion process using empirical conditional statistics further grounds the learnt representations, allowing the model to more accurately capture the underlying risk–return trade-offs in different market states.

\section{Financial Analysis of Portfolios}
\label{app:financial}

To understand the sources of portfolio performance, we conduct standard financial analysis using the Capital Asset Pricing Model (CAPM) and the Fama--French factor model. These models decompose returns into components explained by systematic risk factors and residual alpha, allowing us to assess whether the performance arises from true signal extraction or exposure to common risk premia.

\paragraph{CAPM regression.}
We first estimate the CAPM model on daily returns:
\begin{equation*}
R_{\textsc{radar},t} - R_{f,t} = \alpha + \beta \big(R_{\text{market},t} - R_{f,t}\big) + \epsilon_t,
\end{equation*}
where:
\begin{itemize}[leftmargin=*]
    \item $(R_{\textsc{radar},t} - R_{f,t})$: The excess return of the \textsc{radar} portfolio on day $t$.
    \item $(R_{\text{market},t} - R_{f,t})$: The excess return of the market benchmark on day $t$.
    \item \textbf{Alpha ($\alpha$):} The intercept, representing the portion of returns not explained by market exposure.
    \item \textbf{Beta ($\beta$):} The slope, which measures the \textsc{radar}  portfolio’s systematic risk relative to market.
\end{itemize}
A strong strategy should exhibit a positive and statistically significant $\alpha$, indicating that it generates excess returns beyond what can be explained by market exposure. An $\alpha$ close to zero would suggest performance comparable to the CAPM benchmark. In addition, we report the $R^2$, which measures the proportion of return variation explained by the model. A lower $R^2$ implies that the strategy is not simply replicating common risk exposures, and is instead capturing useful idiosyncratic signals.

\begin{table}[h]
\centering
\small
\caption{CAPM regression results. Significance levels: $^{***}p<0.01$, $^{**}p<0.05$, $^{*}p<0.1$.}
\label{tab:capm}
\begin{tabular}{lccccc}
\toprule
Model & $\alpha$ & $t$-stat & $p$-value & $\beta$ & $R^2$ \\
\midrule
Equal Weight  & 4.78\%*** & 2.712 & 0.003 & 1.030 & 0.957 \\
BSV           & 6.60\%*   & 1.614 & 0.053 & 1.248 & 0.849 \\
DKKM          & 4.10\%*** & 2.403 & 0.008 & 1.019 & 0.960 \\
MLP           & 4.86\%**  & 1.864 & 0.031 & 0.893 & 0.897 \\
LinAttn       & 7.13\%**  & 2.208 & 0.014 & 1.051 & 0.859 \\
SDF           & 5.34\%*   & 1.352 & 0.088 & 0.983 & 0.816 \\
\textsc{radar} (Ours)  & \textbf{18.71\%**} & 2.159 & 0.016 & 0.977 & 0.476 \\
\bottomrule
\end{tabular}
\end{table}

The CAPM results reveal several key findings. Firstly, \textsc{radar} achieves the highest annualized alpha (18.71\%), which is statistically significant at the 5\% level. This indicates that its performance cannot be explained solely by market exposure. In contrast, baseline models exhibit substantially lower alpha, even when statistically significant.

Secondly, the estimated beta for \textsc{radar} is close to one (0.977), suggesting moderate market exposure rather than a market-neutral strategy. Importantly, the relatively low $R^2$ (0.476) indicates that a large portion of return variation is not explained by market movements, pointing to meaningful idiosyncratic signal extraction. By comparison, traditional models such as Equal Weight exhibit much higher $R^2$ values (above 0.95), implying that their returns are largely driven by market factors.

\paragraph{Fama--French factor analysis.}
We further analyze performance using a three-factor model:
\begin{equation*}
R_{p,t} - R_{f,t} = \alpha + \beta_m (R_m - R_f) + \beta_s \text{SMB} + \beta_h \text{HML} + \epsilon_t,
\end{equation*}
where:
\begin{itemize}[leftmargin=*]
    \item $\text{SMB}$: Size factor, defined as the return difference between small-cap and large-cap stocks.
    \item $\text{HML}$: Value factor, defined as the return difference between high and low book-to-market stocks.
    \item $\beta_m, \beta_s, \beta_h$: Factor loadings measuring exposure to the market, size, and value factors, respectively.
\end{itemize}
Similar to the CAPM setting, we look for a positive and statistically significant value $\alpha$, which indicates excess returns beyond common risk factors. The $R^2$ now reports the proportion of return variation explained jointly by all factors. A lower $R^2$ suggests that the strategy captures useful idiosyncratic signals. The $\beta$ coefficients capture the model’s exposure to the respective factors. 
\begin{table}[h]
\centering
\small
\caption{Fama--French regression results. Significance levels: $^{***}p<0.01$, $^{**}p<0.05$, $^{*}p<0.1$.}
\label{tab:ff}
\begin{tabular}{lccccccc}
\toprule
Model & $\alpha$ & $t$-stat & $p$-value & $\beta_m$ & $\beta_{SMB}$ & $\beta_{HML}$ & $R^2$ \\
\midrule
Equal Weight  & 4.63\%*** & 3.233 & 0.001 & 0.999 & 0.079 & 0.159 & 0.975 \\
BSV           & 8.73\%*** & 2.833 & 0.002 & 1.192 & 0.402 & 0.327 & 0.918 \\
DKKM          & 3.88\%*** & 2.735 & 0.003 & 0.988 & 0.042 & 0.156 & 0.974 \\
MLP           & 2.30\%    & 0.870 & 0.192 & 0.892 & -0.079 & -0.064 & 0.896 \\
LinAttn       & 8.05\%*** & 2.448 & 0.007 & 1.007 & 0.134 & 0.245 & 0.889 \\
SDF           & 7.42\%**  & 2.002 & 0.023 & 0.927 & 0.108 & 0.347 & 0.863 \\
\textsc{radar} (Ours)  & \textbf{15.47\%**} & 1.814 & 0.035 & 0.988 & 0.064 & -0.173 & 0.488 \\
\bottomrule
\end{tabular}
\end{table}

Under the multi-factor specification, \textsc{radar} continues to exhibit strong and statistically significant alpha (15.47\%), confirming that its performance cannot be attributed to the other common risk factors such as size or value. Notably, \textsc{radar} shows relatively small exposure to SMB and a negative loading on HML, suggesting that its returns are not driven by standard factor tilts such as small-cap or value strategies. It also achieves the lowest average $R^2$ among all baseline models, indicating that a much smaller fraction of its overall returns variation can be explained by the compared factors. 

Overall, these results demonstrate that \textsc{radar} captures economically meaningful alpha beyond market explanations. The combination of high alpha, moderate beta, and low $R^2$ supports the conclusion that the model extracts genuine predictive signals rather than relying on factor exposures.